\documentclass[iicol,pdflatex]{mysn-jnl}

\usepackage{amsmath,empheq,nicefrac,amssymb,bm,mathtools}

\usepackage{natbib}

\usepackage{multicol}

\usepackage{epsfig}

\usepackage{graphicx}
\usepackage[crop=pdfcrop,mode=batch]{pstool}

\usepackage{booktabs,multirow}

\usepackage{comment}

\usepackage{forloop,siunitx,fmtcount}

\usepackage{balance}

\definecolor{mygreen}{rgb}{0,0.6,0}
\definecolor{mygreen2}{rgb}{0,.55,0}
\definecolor{mygrey}{rgb}{.29,.29,.29}
\definecolor{myblue}{rgb}{.12,.46,.70}
\hypersetup
{colorlinks=true,
	linkcolor=red,
	citecolor=myblue,
	filecolor=magenta,
	urlcolor=myblue,
	linktoc=page,
}

\usepackage{doi}

\usepackage{csquotes}

\usepackage{subcaption}
\newcommand{\vr}[1]{{\mbox{\bm{$#1$}}}}

\newcommand{\trans}{^{\ensuremath{\mathsf{T}}}}

\newcommand{\bigO}{\ensuremath{\mathcal{O}}}

\renewcommand{\x}{\ensuremath{\vr{x}}}
\newcommand{\q}{\ensuremath{\vr{q}}}
\renewcommand{\v}{\ensuremath{\vr{v}}}
\renewcommand{\u}{\ensuremath{\vr{u}}}
\newcommand{\dx}{\ensuremath{\vr{\dot{x}}}}
\newcommand{\dq}{\ensuremath{\vr{\dot{q}}}}
\newcommand{\dv}{\ensuremath{\vr{\dot{v}}}}
\newcommand{\ddq}{\ensuremath{\vr{\ddot{q}}}}

\renewcommand{\f}{\ensuremath{\vr{f}}}
\newcommand{\g}{\ensuremath{\vr{g}}}

\newcommand{\TZ}[1]{\mbox{\emph{Tz}-{\ensuremath{#1}}}}
\newcommand{\HS}[1]{\mbox{\emph{HS}-{\ensuremath{#1}}}}

\begin{document}

\title[\qquad]{
		Collocation methods for second and higher order systems
}

\author*{\fnm{Siro} \sur{Moreno-Mart\'{\i}n$^{*}$}}\email{smorenom@iri.upc.edu}
\author{\fnm{Llu\'{\i}s} \sur{Ros}}\email{ros@iri.upc.edu}
\author{\fnm{Enric} \sur{Celaya}}\email{ecelaya@iri.upc.edu}

\affil{\orgdiv{\centering Institut de Rob\`otica i Inform\`atica Industrial (CSIC-UPC)\\} \orgaddress{\street{Llorens Artigas 4-6}, \city{Barcelona}, \postcode{08028}, \state{Catalonia}, \country{Spain}}}


\abstract{It is often unnoticed that the predominant way to use collocation methods is fundamentally flawed when applied to  optimal control in robotics. 
Such methods assume that the system dynamics is given by a first order ODE, whereas robots are often governed by a second or higher order ODE involving configuration variables and their time derivatives. To apply a collocation method, therefore, the usual practice is to resort to the well known procedure of casting an \emph{M}th order ODE into \emph{M} first order ones. This manipulation, which in the continuous domain is perfectly valid, leads to inconsistencies when the problem is discretized. Since the configuration variables and their time derivatives are approximated with polynomials of the same degree, their differential dependencies cannot be fulfilled, and the actual dynamics is not satisfied, not even at the collocation points. This paper draws attention to this problem, and develops improved versions of the trapezoidal and Hermite-Simpson collocation methods that do not present these inconsistencies. In many cases, the new methods reduce the dynamic transcription error in one order of magnitude, or even more, without noticeably increasing the cost of computing the solutions.}


\maketitle

\section{Introduction} 
\label{sec:introduction}

\vspace*{1.1mm}

Direct collocation methods have proven to be powerful tools for solving optimal control problems in robotics \citep{posa2016optimization,pardo2016evaluating,kelly2017introduction,hereid2018dynamic,tedrake2023underactuated}. Initially developed for aeronautics and astrodynamics applications \citep{hargraves1987direct,conway2010spacecraft}, these methods have become very popular and of widespread use in the context of trajectory optimization and model predictive control, thanks to a few key advantages over indirect approaches based on the Pontryagin conditions of optimality: in general, they are easier to implement and show larger regions of convergence, and do not require estimations of the costate variables, which may be difficult to obtain accurately. Helpful tutorials and monographs like \cite{kelly2017introduction} or \cite{betts2010practical}, as well as open-source software for nonlinear optimization \citep{wachter2006ipopt}, numerical optimal control~\citep{kelly2017introduction,becerra2010psopt,andersson2019casadi}, or model-based design and verification~\citep{drake}, are also contributing to their rapid dissemination among the community.

Direct collocation methods involve the transcription of the continuous-time optimal control problem into a finite-dimensional nonlinear programming (NLP) problem~\citep{kelly2017introduction}. The transcription is based on partitioning the time history of the control and state variables into a number of intervals delimited by knot points. The system dynamics is then discretized in each interval by imposing the differential constraints at a set of collocation points, which may coincide, or not, with the chosen knot points. The cost function is also approximated using the values taken by the variables at such points, and the NLP problem is formulated using them. Once this problem is solved, a continuous solution is built using interpolating polynomials that satisfy the dynamics equations at the collocation points. 

The general formulation of most collocation methods assumes that the system dynamics is governed by a first order ODE of the form
\begin{equation} \label{eq:f}
	\dx(t) = \vr{f}(\x(t), \u(t), t),
\end{equation}
where $\vr{x}(t)$ and $\vr{u}(t)$ are the state trajectory and the control function, respectively \citep{tedrake2023underactuated}. In robotics, however, as in mechanics in general, the evolution of the system is often determined by a second order ODE of the form
\begin{equation}\label{eq:g}
	\ddq(t) = \g \left( \q(t),\dq(t),\u(t),t \right),
\end{equation}
where $\vr{q}(t)$ is the configuration trajectory and $\dq(t)$ is its time derivative. To apply a general collocation method, therefore, the usual procedure is to define $\v(t) = \dq(t)$ and write \eqref{eq:g} as 
\begin{subequations}
	\label{eq:gv}
	\begin{empheq}[left=\empheqlbrace]{align}
		\dq(t) &= \v(t), \\
		\dv(t) &= \g(\q(t),\v(t),\u(t),t).
	\end{empheq}
\end{subequations}
which, if we define $\x(t) = (\q(t),\v(t))$, corresponds formally to \eqref{eq:f}.
Yet, this raises a consistency issue. 
Since the collocation method locally approximates $\q(t)$ and $\v(t)$ by polynomials of the same degree, imposing  
\begin{align}
	\vr{v}(t)=\dq(t) \label{eq:functrel}
\end{align}
only at the collocation points does not grant the satisfaction of \eqref{eq:functrel} over the continuous time domain. Even more striking, perhaps, is the fact that, as we demonstrate in this paper, imposing \eqref{eq:gv} at the collocation points does not imply the satisfaction of \eqref{eq:g}, not even at these points,
which contributes to increase the dynamic transcription error along the obtained trajectories. This hinders the possibility to reach a correct solution since, even if $\vr{u}(t)$ produces the expected trajectory for $\vr{v}(t)$, its integration will rarely coincide with the function obtained for $\vr{q}(t)$. In other words, the state trajectory $\vr{x}(t)$ will be inconsistent in general.  

In this work, we present modified versions of the trapezoidal and Hermite-Simpson collocation methods specifically addressed to solve these issues for second order systems with the dynamics in \eqref{eq:g}. The new formulations grant that the collocation polynomials fulfill the condition in \eqref{eq:functrel} while satisfying $\eqref{eq:g}$ at the collocation points, thereby increasing the accuracy of the obtained solutions. We early presented these methods in \citet{moreno2022collocation}, and here we also extend them to treat $M$th order ODEs of the form
\begin{equation}\label{eq:g_M}
	\scalebox{0.93}{$\q^{(M)}(t) = \g(\q(t),\dq(t),...,\q^{(M-1)}(t),\u(t),t)$},
\end{equation}
where $\q^{(M)}(t)=d^M\q(t)/dt^M$. While less common, ODEs of this kind arise more or less explicitly in flexible, elastic, or soft robots for example \citep{deluca2016robots,dellasantina2020flexible}.

By means of well-established benchmark problems from the literature, we further demonstrate that the new methods reduce substantially the dynamic transcription error (in one order of magnitude or even more depending on the number of knot points) without noticeably increasing the computational time needed to solve the transcribed NLP problems. As a result, the state and control trajectories $\x(t)$ and $\u(t)$ will be mutually more consistent, thus facilitating their tracking with a feedback controller.

The rest of the paper is structured as follows. Section~\ref{sec:formulation} formulates the optimal control problem to be solved and delimits the specific transcription problem that we face in this paper. To prepare the ground for later developments, Section~\ref{sec:first_order} reviews the conventional trapezoidal and Hermite-Simpson collocation methods for first order systems and pinpoints their limitations on transcribing second order differential equations. Improved versions of these methods are developed in Section~\ref{sec:second_order} for second order systems, and for $M$th order ones in Section~\ref{sec:higher_order}. The methods are summarized and compared in Section \ref{sec:summary}, and their performance is analysed in Section \ref{sec:tests} with the help of examples. Finally, Section~\ref{sec:conclusions} concludes the paper and enumerates a few points for future attention.

\section{Problem formulation}
\label{sec:formulation}

The optimal control problem that concerns us in this paper consists of finding state and action trajectories $\vr{x}(t)$ and $\vr{u}(t)$, and a final time $t_f$, that
\begin{subequations}
	\label{eq:OCP}
	\begin{align}
		&\text{minimize} \hspace{5cm} \notag \\
		\label{eq:OCP_cost}
		& \qquad K(\x_f,t_f) + \int_{0}^{t_f}L(\x(t),\u(t))\;dt \\
		&\text{subject to} \hspace{5cm} \notag \\
  	\label{eq:OCP_dynamics}
		& \qquad \dx(t) = \f(\x(t),\u(t),t), \hspace{6.3mm} t \in [0,t_f] 
        \\
		\label{eq:OCP_path}
		& \qquad \vr{p}(\x(t),\u(t)) \le \vr{0}, \hspace{13.5mm} t \in [0,t_f] 
        \\
		\label{eq:OCP_boundary}
		& \qquad \vr{b}(\x_0, \x_f,t_f) = \vr{0}, 
        \\
		& \qquad t_f \geq 0,
	\end{align}
\end{subequations}
where $K(\x_f,t_f)$ and $L(\x(t),\u(t))$ are terminal and running cost functions, respectively, Eq.~\eqref{eq:OCP_dynamics} is an ODE modeling the system dynamics, Eqs.~\eqref{eq:OCP_path} and~\eqref{eq:OCP_boundary} encompass the path and boundary constraints, $\x_0 = \x(0)$, and $\x_f = \x(t_f)$. 

We note that, while Eq.~\eqref{eq:OCP_dynamics} has the appearance of a first order ODE, in robotics it often takes the form
\begin{align}
	\left.
	\begin{aligned}
		& \dx_1 = \x_2           \\
		& \dx_2 = \x_3           \\
		& \hspace{3mm} \vdots    \\
		& \dx_{M-1} = \x_M       \\
		& \dx_M = \g(\x,\u,t) 
		\hspace{2mm}
	\end{aligned} 
	\right\}
\end{align}
where 
\begin{equation}
	\x = (\x_1,\ldots,\x_M) = (\q,\dq,\ldots,\q^{M-1}),
\end{equation}
so in such cases it actually encodes an $M$th order ODE like \eqref{eq:g_M}, or \eqref{eq:g} if $M=2$.

Solving Problem~\eqref{eq:OCP} via collocation involves partitioning the time history of the control and state variables into $N$ intervals delimited by $N+1$ knot points $t_k$, $k=0, \ldots, N$, then transcribing Eqs. \eqref{eq:OCP_cost}-\eqref{eq:OCP_path} into appropriate discretizations expressed in terms of the values $\vr{x}_k = \vr{x}(t_k)$ and $\vr{u}_k = \vr{u}(t_k)$, and finally solving the constrained optimization problem that results. 

The transcriptions of \eqref{eq:OCP_cost} and \eqref{eq:OCP_path} are relatively straightforward and less relevant in the context of this paper. They can be done, for example, by approximating the integral in \eqref{eq:OCP_cost} using some quadrature rule, and enforcing \eqref{eq:OCP_path} for all knot points $t_k$. The transcription of \eqref{eq:OCP_dynamics}, in contrast, is substantially more involved, and will be the main subject of the rest of this paper. In particular, we seek to construct appropriate polynomial approximations of the solutions $\vr{x}(t)$ of \eqref{eq:OCP_dynamics} for each interval $[t_k,t_{k+1}]$. These approximations will be defined as solutions of systems of equations which, when considered together for all intervals, will form a proper transcription of \eqref{eq:OCP_dynamics} over the whole time domain $[0,t_f]$.

\section{Methods for first order systems}
\label{sec:first_order}

Two of the most widely used transcriptions of \eqref{eq:OCP_dynamics} are those of the trapezoidal and Hermite-Simpson methods, which assume no particular form for~\eqref{eq:OCP_dynamics}. To see where these transcriptions incur in dynamical error, and ease the development of the new methods, we briefly explain how they approximate \eqref{eq:OCP_dynamics} and obtain their approximation polynomials for the state. Our results match those by \cite{betts2010practical} and \cite{kelly2017introduction}, but we follow a derivation process that is closer to \citet{hargraves1987direct}, which facilitates the transition to our new methods in Sections~\ref{sec:second_order} and~\ref{sec:higher_order}.

\subsection{Trapezoidal collocation}
\label{subsec:trap1}

In trapezoidal collocation, the state trajectories are approximated by quadratic polynomials. If, for each interval $[t_k, t_{k+1}]$, we define $\tau = t-t_k$, we can write the polynomial approximation for a component $x$ of the state in this interval, and its temporal derivative, as
\begin{subequations} \label{eq:tzpoly}
	\begin{align}
		x(t) &= a  + b \tau + c\tau^2,  \label{eq:tzpolyx} \\
		\dot{x}(t) &=   b + 2c \tau,    \label{eq:tzpolyxdot}
	\end{align}
\end{subequations}
where $a$, $b$, and $c$ are real coefficients. To facilitate the application of collocation constraints, however, we will rewrite $x(t)$ using the three parameters
\begin{align}
	x_k &= x(t_k), \label{eq:tzpar1} \\ 
	\dot{x}_k &= \dot{x}(t_k), \label{eq:tzpar2} \\
	\dot{x}_{k+1} &= \dot{x}(t_{k+1}). \label{eq:tzpar3}
\end{align}
Evaluating the right-hand sides of \eqref{eq:tzpar1}-\eqref{eq:tzpar3} using~\eqref{eq:tzpoly} we obtain
\begin{equation}
	\begin{bmatrix}
		x_k \\
		\dot{x}_k \\ 
		\dot{x}_{k+1}
	\end{bmatrix}
	=
	\begin{bmatrix}
		1 & 0 & 0\\
		0 & 1 & 0 \\ 
		0 & 1 & 2h
	\end{bmatrix}
	\begin{bmatrix}
		a \\
		b \\ 
		c
	\end{bmatrix},
\end{equation}
where $h=t_{k+1}-t_k$, so solving for $a,b,c$ and substituting the resulting expressions in \eqref{eq:tzpolyx}, we have
\begin{equation}\label{eq:tzinterpol}
x(t) = x_k + \dot{x}_k \tau+\dfrac{\tau^2}{2h}(\dot{x}_{k+1}-\dot{x}_k).
\end{equation}
Equation \eqref{eq:tzinterpol} is known as the interpolation polynomial, as it allows us to estimate the intermediate states for $t \in [t_k,t_{k+1}]$, once the NLP problem has been solved.

Now, following \citet[page 30]{hairer2002geometric}, we determine the three parameters of \eqref{eq:tzinterpol} by enforcing the initial value constraint $x(t_k)=x_k$ and two collocation constraints of the form 
\begin{equation*}
\dot{x}(t_i) = f(\vr{x}(t_i),\vr{u}(t_i),t_i).
\end{equation*}
From \eqref{eq:tzinterpol} we see that $x(t_k)=x_k$ by construction. As for the collocation constraints, the trapezoidal method imposes them at the knot points $t_k$ and $t_{k+1}$, so it must be
\begin{align}
\dot{x}_k &= f_k, \label{eq:tzcol1} \\
\dot{x}_{k+1} &= f_{k+1}, \label{eq:tzcol2} 
\end{align}
where $f_k$ is a shorthand for $f({\vr{x}}_k, {\vr{u}}_k, t_k)$. The value $x_{k+1}$, then, is obtained by evaluating \eqref{eq:tzinterpol} for $\tau = h$. This results in the constraint
\begin{equation}\label{eq:tzfinval}
{x}_{k+1} = {x_k} + \dfrac{h}{2}(\dot{x}_{k+1} + \dot{x}_k),
\end{equation}
which ensures the continuity of the trajectory across intervals $k$ and $k+1$.

Note that equations \eqref{eq:tzcol1}-\eqref{eq:tzfinval} already form a transcription of our ODE in the interval $[t_k,t_{k+1}]$ since, if $\vr{x}_k$, $\vr{u}_k$, and $\vr{u}_{k+1}$ were known, these equations would suffice to determine the three unknowns $\dot{x}_k$, $\dot{x}_{k+1}$, and $x_{k+1}$. However, we can also substitute \eqref{eq:tzcol1} and \eqref{eq:tzcol2} into \eqref{eq:tzfinval} to obtain the more compact expression
\begin{equation}\label{eq:tzcompressed}
{x}_{k+1} = {x_k} + \dfrac{h}{2}(f_{k+1} + f_k),
\end{equation}
which we recognize as the common transcription rule in trapezoidal collocation \citep{kelly2017introduction,betts2010practical}. Observe that the continuity between the polynomials of intervals $k$ and $k+1$ is granted for the first derivative as, by construction, they both satisfy $\dot{x}_{k+1} = f_{k+1}$. However, second and higher order continuity is not preserved in general.

\subsection{Hermite-Simpson collocation}
\label{subsec:hs1}

In Hermite-Simpson collocation, the state trajectories in each interval are approximated by cubic polynomials:
\begin{subequations} \label{eq:HSpoly}
\begin{align}
	x(t) &= a + b \tau + c \tau^2 +d \tau^3 , \label{eq:HSpolyx} \\
	\dot{x}(t) &= b + 2c\tau + 3d\tau^2. \label{eq:HSdpolyxdot}
\end{align}
\end{subequations}
By analogy with the trapezoidal method, we first express the polynomial coefficients in terms of the parameters
\begin{align*}
x_k &= x(t_k),\\ 
\dot{x}_{k} &= \dot{x}(t_k),\\
\dot{x}_{c} &= \dot{x}(t_c),\\
\dot{x}_{k+1} &= \dot{x}(t_{k+1}),
\end{align*}
where $t_c = t_k + h/2$, and the extra parameter $\dot{x}_c$ is added because four parameters are needed to determine a third degree polynomial. Evaluating these identities using \eqref{eq:HSpoly}, solving for $a,\ldots,d$, and substituting the expressions in \eqref{eq:HSpolyx}, we obtain the interpolation polynomial
\begin{align}
\begin{split} \label{eq:HSinterpol}
	x(t) = x_k & + \dot{x}_k \tau 
	- \dfrac{\tau^2}{2h}(3\dot{x}_k - 4\dot{x}_c + \dot{x}_{k+1}) + \\
	& + \dfrac{\tau^3}{3h^2}(2\dot{x}_k-4\dot{x}_c + 2\dot{x}_{k+1}).	
\end{split}
\end{align}

In order to determine the four parameters of \eqref{eq:HSinterpol}, four conditions have to be imposed, and the Hermite-Simpson method makes this by fixing $x(t_k) = x_k$ (which holds by construction) and imposing the dynamics at the two bounding knot points and the midpoint between them:
\begin{align}
\dot{x}_k &= f_k, \label{eq:HScol1} \\
\dot{x}_{k+1} &= f_{k+1}, \label{eq:HScol2} \\
\dot{x}_{c} &= f_{c}. \label{eq:HScol3} 
\end{align}
In the latter equation, $f_c = f(\vr{x}_c,\vr{u}_c,t_c)$, where $\vr{x}_c=\vr{x}(t_c)$, and $\vr{u}_c=\vr{u}(t_c)$. Moreover, the values $x_c$ that are needed in $f_c$ can be expressed in terms of the above four parameters by evaluating \eqref{eq:HSinterpol} for $\tau = h/2$, which yields
\begin{align} 
\label{eq:HSxc}
x_c = x_k + \frac{h}{24}(5\dot{x}_k + 8\dot{x}_c - \dot{x}_{k+1}).
\end{align}
Finally, the continuity constraint between intervals $k$ and $k+1$ is obtained by evaluating \eqref{eq:HSinterpol} for $\tau = h$:
\begin{align} 
\label{eq:HSxkp1}
x_{k+1} &= x_k + \frac{h}{6}(\dot{x}_k + 4\dot{x}_c + \dot{x}_{k+1}).
\end{align}

Equations \eqref{eq:HScol1}-\eqref{eq:HSxkp1} already form a transcription of our ODE in $[t_k,t_{k+1}]$, but a transcription involving less variables can be obtained by substituting \eqref{eq:HScol1}-\eqref{eq:HScol3} in \eqref{eq:HSxkp1} and \eqref{eq:HSxc}, which gives
\begin{subequations}
\label{eq:HSsep1}
\begin{align}
	&x_{k+1} = x_k + \frac{h}{6}(f_k + 4f_c + f_{k+1}), \label{eq:HSxkp1alt} \\
	&x_c = x_k + \frac{h}{24}(5f_k + 8f_c - f_{k+1}). \label{eq:HSxcalt}
\end{align}
\end{subequations}

If preferred, we can also remove the dependence on $f_c$ in \eqref{eq:HSxcalt}. This is achieved by isolating $f_c$ from \eqref{eq:HSxkp1alt} and substituting the result in \eqref{eq:HSxcalt}, which yields the alternative transcription
\begin{subequations} 
\label{eq:HSsep2}
\begin{align}
	&x_{k+1} = x_k + \frac{h}{6}(f_k + 4f_c + f_{k+1}), \label{eq:HSxkp1alt_repe} \\
	&x_c = \frac{1}{2}(x_k + x_{k+1}) + \frac{h}{8}(f_k - f_{k+1}). \label{eq:HSxckelly}
\end{align}
\end{subequations}

Both transcriptions in \eqref{eq:HSsep1} and \eqref{eq:HSsep2} are called separated forms of Hermite-Simpson collocation, in the sense they both keep $x_c$ as a decision variable of the problem. They are equivalent, but the one in \eqref{eq:HSsep2} allows us to eliminate $x_c$ by substituting \eqref{eq:HSxckelly} in \eqref{eq:HSxkp1alt_repe}, which results in a single equation that is known as the compressed form of Hermite-Simpson collocation~\citep{kelly2017introduction,betts2010practical}. While the use of a separated form tends to be better when working with a small number of intervals, the compressed form is preferable when such a number is large~\citep{kelly2017introduction}.

Note that, despite the polynomial approximation into each interval between consecutive knot points is of third degree, continuity through knot points is only granted for the state trajectory and its first derivative.

\subsection{Trajectory interpolation}
\label{subsec:interpolation}

After solving the NLP problem, values of the state and control variables at all collocation points are available. A continuous approximation to the optimal trajectory for the state is then obtained by substituting \eqref{eq:tzcol1}-\eqref{eq:tzcol2}, or \eqref{eq:HScol1}-\eqref{eq:HScol3}, in the corresponding interpolating polynomials \eqref{eq:tzinterpol} and \eqref{eq:HSinterpol}, for the trapezoidal and Hermite-Simpson methods, respectively. The approximation of the control trajectory within each interval is obtained, in the trapezoidal case, by linear interpolation of the control values. In the Hermite-Simpson case, different options are possible. 
Some authors handle the midpoint control as an independent variable and use a quadratic interpolation of the three control values available in each interval \citep{kelly2017introduction}, while others prefer a linear interpolation and enforce the midpoint value to be the mean of the two bounding values \citep{topputo2014survey}. In this paper we follow the former option.

\subsection{Downsides of the methods}
\label{subsec:drawbacks}

\begin{figure}[t!]
    \begin{center}
	   \includegraphics[width=1\linewidth]{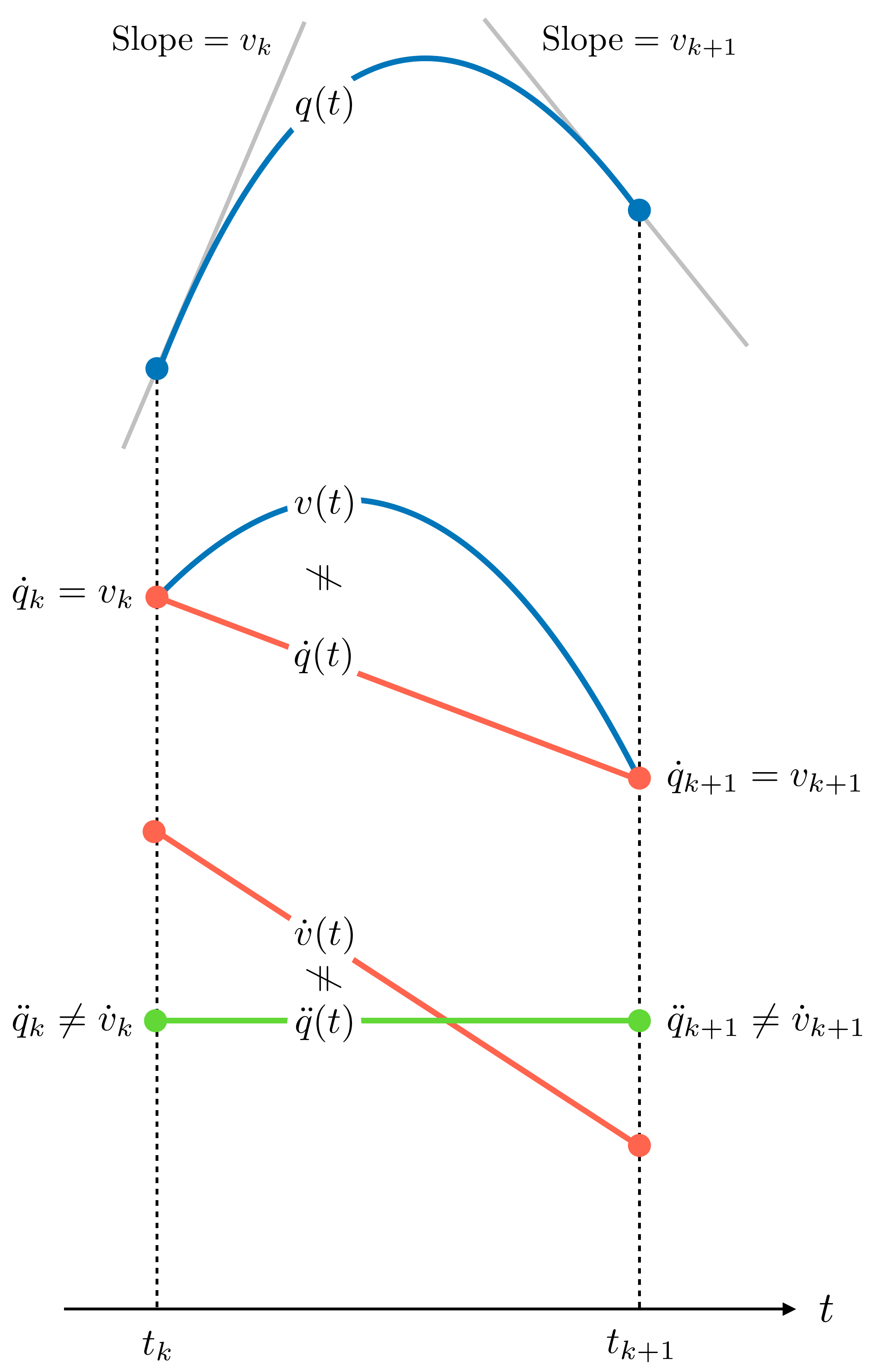}
    \end{center}
    \caption{\label{fig:inconsistencies}Inconsistencies that arise when a collocation method for first order systems is applied to a second order ODE $\ddq = \vr{g}(\vr{q},\dq,\vr{u},t)$. The figure illustrates the case of the trapezoidal method, whose quadratic approximations $q(t)$ and $v(t)$ are depicted in blue. The red and green curves correspond to first and second derivatives of these trajectories, respectively.}
\end{figure}

In a first order dynamical system, imposing \eqref{eq:tzcol1}-\eqref{eq:tzcol2} or \eqref{eq:HScol1}-\eqref{eq:HScol3} grants that the system dynamics is effectively satisfied at the collocation points. The same is not true when a second order system is cast into a first order one via \eqref{eq:gv}. To see why, note that the constraint $\dot q(t) = v(t)$ is only imposed at the collocation points, but not in between them, so that, even if the curves $\dot q(t)$ and $v(t)$ coincide at such points, their derivatives may be different in them (Fig.~\ref{fig:inconsistencies}).
Therefore, $\ddot q(t) \neq \dot v(t)$ in general and, in particular, also at the collocation points. 
As a consequence, even if $\dot q_k=v_k$ and $\dot v_k=g(\vr{q}_k,\vr{v}_k,\vr{u}_k,t_k)$, this does not imply that the expected relation $\ddot{q}_k = g(\q_k,\v_k,\u_k,t_k)$ is satisfied, what means that, with a transcription based on \eqref{eq:gv}, the system dynamics in \eqref{eq:g} is not granted, not even at the collocation points.
This problem is solved in the second order collocation methods introduced in the next section.

A related problem of first order methods is that, when the trajectories are approximated with their interpolation polynomials $q(t)$ and $v(t)$, the difference $v(t) - \dot{q}(t) \neq 0$ makes the state trajectory inconsistent, so that, if we try to follow it with a controller, since the configuration and velocity trajectories are incompatible, both cannot be followed at the same time. An attempt to solve this may consist in ignoring the configuration trajectory and replacing it by the integral of the velocity, but the resulting configuration trajectory may violate the problem constraints, e.g., the final configuration may be different from the expected one. Alternatively, one can try to replace the velocity trajectory by the derivative of $q(t)$, but in this case, since the dynamic constraint satisfied at collocation point $k$ is $\dot{v}_k=g(\vr{q}_k, \vr{v}_k, \vr{u}_k, t_k)$, and  ${v}_k = \dot{q}_k$ but $\dot{v}_k \neq \ddot{q}_k$, the dynamic constraint $\ddot{q}_k=g(\vr{q}_k, \vr{\dot{q}}_k,\vr{u}_k, t_k)$ will not be satisfied with the computed $\vr{u}_k$.


\section{Methods for second order systems}
\label{sec:second_order}

To solve the inconsistency problems just explained, we propose alternative formulations for the trapezoidal and Hermite-Simpson collocation methods in which the dynamic constraints are directly imposed on the second derivative of the configuration variables, instead of on the first derivative of the state variables. 
By doing so, the velocity variables are not treated as independent from the configuration ones,
but explicitly defined as $v(t) \equiv \dot{q}(t)$. In this way, the discrepancy between $q(t)$ and $v(t)$ is fully removed, and the second order dynamics is satisfied at each collocation point.

\subsection{Trapezoidal method for second order systems}
\label{subsec:trap2}

The essential feature characterizing trapezoidal collocation is that the dynamics is imposed just at the knot points or, otherwise said, that each interval bound is a collocation point.
When the dynamics is governed by the second order ODE in \eqref{eq:g},
using the same strategy as the trapezoidal method consists in imposing \eqref{eq:g} at each interval bound. 
This means that, for each interval, two constraints have to be imposed  on the second derivative of the polynomial approximating each component $q$ of the configuration. But, since the second derivative of a quadratic polynomial is constant, only one constraint could be imposed on it. This implies that the interpolating polynomial $q(t)$ must be of degree three at least. So, we will have, for a given interval $[t_k,t_{k+1}]$,
\begin{subequations}
\label{eq:TZ2poly}
\begin{align}
	q(t) &= a  + b \tau + c\tau^2 +d\tau^3, \label{eq:TZ2poly1} \\
	\dot{q}(t) &= b + 2c\tau + 3d\tau^2, \label{eq:TZ2poly2}   \\
	\ddot{q}(t) &=2c + 6d\tau.\label{eq:TZ2poly3} 
\end{align}
\end{subequations}

To determine the coefficients $a, b, c, d$, we need to impose four conditions. While in the trapezoidal method three conditions were used (the value $x_k$ at the initial bound and the derivatives $\dot{x}_k$ and $\dot{x}_{k+1}$ at the two bounds), here we will impose, in addition to the initial value $q_k$ and the second derivative at the interval bounds $\ddot{q}_k$ and $\ddot{q}_{k+1}$, the value $\dot{q}_k$ of the first derivative at the initial bound.
Note that, for a cubic polynomial, no more than two independent conditions can be fulfilled by its second derivative, so imposing the dynamics at the midpoint of the interval as in the Hermite-Simpson method is not possible here. Thus we will use as parameters:
\vspace*{-2mm}
\begin{align*}
q_k &= q(t_k) \\ 
\dot{q}_k &= \dot{q}(t_k) \\
\ddot{q}_k &= \ddot{q}(t_k) \\
\ddot{q}_{k+1} &= \ddot{q}(t_{k+1}).
\end{align*}
Evaluating these identities using \eqref{eq:TZ2poly} and solving for $a,b,c,d$, we can write the interpolation polynomial $q(t)$ as:
\vspace*{-2mm}
\begin{equation}\label{eq:TZ2interpol}
q(t) = q_k + \dot{q}_k \tau +\ddot{q}_k\dfrac{\tau^2}{2}+ \dfrac{\tau^3}{6h}(\ddot{q}_{k+1}-\ddot{q}_k).
\end{equation}
The evaluation of this polynomial and its derivative $\dot{q}(t)$ for $\tau=h$ yields
\vspace*{-2mm}
\begin{subequations}
\label{eq:TZ2k+1}
\begin{align}
	q_{k+1} &= {q_k}+\dot{q}_k h+\dfrac{h^2}{6}(\ddot{q}_{k+1}+2\ddot{q}_k) \label{eq:TZ2qk+1}, \\
	\dot{q}_{k+1} &= \dot{q}_k + \dfrac{h}{2}(\ddot{q}_{k+1}+\ddot{q}_k) \label{eq:TZ2vk+1},
\end{align}
\end{subequations}
and imposing the collocation constraints
\vspace*{-2mm}
\begin{subequations}
\label{eq:tz2col}
\begin{align}
	\ddot{q}_k &= g_k, \label{eq:tz2col1} \\ 
	\ddot{q}_{k+1} &= g_{k+1}, \label{eq:tz2col2}
\end{align}
\end{subequations}
where $g_k = g(\q_k, \dq_k, \u_k,t_k)$, we finally obtain the trapezoidal method for second order systems:
\vspace*{-2mm}
\begin{subequations}
\label{eq:TZtrans}
\begin{align}
	{q}_{k+1} &= {q_k} + \dot{q}_k h+\dfrac{h^2}{6}(g_{k+1}+ 2g_k), \label{eq:TZtrans_qk+1}\\
	\dot{q}_{k+1} &= \dot{q}_k + \dfrac{h}{2}(g_{k+1}+g_k) \label{eq:TZtrans_vk+1}.
\end{align}
\end{subequations}
Note that, in this case, the trapezoidal rule only applies for the velocity, but not for the configuration itself, which is given by equation \eqref{eq:TZtrans_qk+1}.

As opposed to the trapezoidal method for first order systems, the continuity between neighboring polynomials at the knot points is of second order in this case, since the collocation constraints impose the coincidence of the second derivative of $q(t)$.
Second order continuity for the configuration trajectory implies smooth velocity profiles and continuous accelerations, which are desirable properties in many robotics applications \citep{Constantinescu2000smooth, macfarlane2003jerk,berscheid2021jerklimited}.

\subsection{Hermite-Simpson method for second order systems}
\label{subsec:hs2}

Our purpose now is to impose the second order dynamics on the two bounds and the midpoint of each interval, in similarity with the conventional Hermite-Simpson method. Clearly, if we want to impose three conditions to the second derivative of a polynomial $q(t)$, such a derivative must be quadratic at least, what implies that the polynomial must have degree four at least. Thus, we propose to approximate the configuration trajectory, and its derivatives, by
\begin{align} \label{HS2poly}
q(t) &= a  + b \tau + c\tau^2 + d\tau^3 + e\tau^4, \\
\dot{q}(t) &= b+ 2c\tau  + 3d\tau^2 + 4e\tau^3, \label{HS2poly2} \\
\ddot{q}(t) &=  2c+ 6d\tau + 12e\tau^2.
\end{align}
Since five parameters are needed to determine the five coefficients of $q(t)$, we will use, in addition to the three accelerations $\ddot{q}_k,\ddot{q}_c, \ddot{q}_{k+1}$, the values of the configuration coordinate $q_k$ and its derivative $\dot{q}_k$ at the initial point:
\begin{align*}
&q_k = q(t_k) \\ 
&\dot{q}_k = \dot{q}(t_k) \\
&\ddot{q}_k = \ddot{q}(t_k) \\
&\ddot{q}_c = \ddot{q}(t_c) \\
&\ddot{q}_{k+1} = \ddot{q}(t_{k+1}).
\end{align*}
Solving for the coefficients $a,\ldots,e$, we obtain the following expression for the interpolating polynomial:
\begin{align}
\begin{split} \label{eq:HS2interpol}
	q(t) = q_k &+ \dot{q}_k \tau +\dfrac{\tau^2}{2}\ddot{q}_k - \\
	&- \dfrac{\tau^3}{6h}(3\ddot{q}_k - 4\ddot{q}_c + \ddot{q}_{k+1}) + \\
	&+ \dfrac{\tau^4}{6h^2}(\ddot{q}_k-2\ddot{q}_c+\ddot{q}_{k+1}).
\end{split}
\end{align}
Evaluating \eqref{eq:HS2interpol} and its derivative for the value $\tau=h$ results in
\begin{subequations}
\label{eq:HS2k+1}
\begin{align}
	{q}_{k+1} &= {q_k} + \dot{q}_k h + \frac{h^2}{6}(\ddot{q}_{k} + 2\ddot{q}_c), 
	\label{eq:HS2qk+1} 
	\\
	\dot{q}_{k+1} &=  \dot{q}_k + \frac{h}{6}( \ddot{q}_k+ 4\ddot{q}_c + \ddot{q}_{k+1}),
	\label{eq:HS2vk+1}
\end{align}
\end{subequations}
and imposing the collocation constraints
\begin{subequations}
\label{eq:HS2coll}
\begin{align} 
	\ddot{q}_k &= g_k, \label{HS2coll1} \\
	\ddot{q}_c &= g_c, \label{HS2coll2} \\
	\ddot{q}_{k+1} &= g_{k+1}, \label{HS2coll3}
\end{align}
\end{subequations}
yields
\begin{subequations}
\label{eq:HS2cont}
\begin{align}
	{q}_{k+1} &= {q_k} + \dot{q}_k h+\dfrac{h^2}{6}(g_{k}+ 2g_c), 
	\label{eq:HS2cont_q}
	\\
	\dot{q}_{k+1} &=  \dot{q}_k + \frac{h}{6}( g_k+ 4g_c + g_{k+1}),
	\label{eq:HS2cont_v}
\end{align}
\end{subequations}%
where we recognize that \eqref{eq:HS2cont_v} is the Simpson quadrature for the velocity. The terms $g_c$ in  these equations involve the midpoint coordinate \mbox{$q_c = q(t_c)$}, and the velocity $\dot{q}_c = \dot{q}(t_c)$, but these can be obtained by evaluating \eqref{eq:HS2interpol} and its derivative for $\tau=h/2$, and imposing \eqref{eq:HS2coll}, which yields
\begin{subequations}
\label{eq:HS2c}
\begin{align}
	q_c &= q_k + \frac{h}{2}\dot{q}_k + \frac{h^2}{96}(7g_k + 6g_c - g_{k+1}), \label{eq:HS2qc} \\
	\dot{q}_c &= \dot{q}_k + \dfrac{h}{24}(5g_k + 8g_c - g_{k+1}). \label{eq:HS2vc}
\end{align}
\end{subequations}
Note however that, since $q_c$ and $\dot{q}_c$ are to be used in the evaluation of $g_c$, we may prefer not to express them in terms of $g_c$ itself. For this we simply isolate $g_c$ from \eqref{eq:HS2cont_v} and substitute the result in \eqref{eq:HS2c} to obtain:
\begin{subequations}
\label{HS2c_alt}
\begin{align}
	\begin{split}
		&q_c = q_k + \frac{h}{32}(13 \dot{q}_k+3 \dot{q}_{k+1}) + \\
		& \hspace{2cm} + \frac{h^2}{192}(11 g_k-5 g_{k+1}),
	\end{split}
	\\
	&\dot{q}_c = \frac{1}{2}(\dot{q}_k+\dot{q}_{k+1})+\frac{h}{8}(g_k-g_{k+1}).
\end{align}
\end{subequations}
Equations~\eqref{eq:HS2cont} and \eqref{HS2c_alt} together constitute a separated form of the Hermite-Simpson method for $2$nd order systems. Written in this way, \eqref{HS2c_alt} can be replaced in the expression of $g_c$ in \eqref{eq:HS2cont} to transcribe the problem in compressed form, which eliminates the need to treat $q_c$ and $\dot{q}_c$ as decision variables. 

In this collocation scheme, the continuity across knot points is also of second order due to the coincidence of the second derivative imposed by the collocation constraints, what gives rise to smooth, continuous acceleration trajectories just like in the second order trapezoidal method.

\section{Extensions for higher order systems}
\label{sec:higher_order}

Second order systems are, by far, the most common in robotics, but sometimes it may be necessary to deal with dynamical systems of a higher order $M$, whose dynamics is described by an ODE like \eqref{eq:g_M}, which we recall for convenience:
\begin{equation}
	\label{eq:g_high}
	\scalebox{0.9}{$\q^{(M)}(t) = \g\left(\,\q(t),\dq(t),...,\q^{(M-1)}(t),\u(t),t\,\right)$}.
\end{equation}
We next see how the new methods can be extended to transcribe \eqref{eq:g_high}.

\subsection{The generalized trapezoidal method}
\label{subsec:trap_high}

To derive the trapezoidal method for $M$th order systems we proceed as in Section \ref{subsec:trap2}. For each time interval $[t_k,t_{k+1}]$ we approximate each component $q$ of the solution of \eqref{eq:g_high} by a polynomial $q(t)$ whose $M$th time derivative $q^{(M)}(t)$ is linear, so its two coefficients may be determined by imposing \eqref{eq:g_high} at each interval bound. This implies that $q(t)$ must be of order $M+1$. Using $a_0,\ldots,a_{M+1}$ as coefficients, this polynomial, and its derivatives, may be written as
\begin{subequations}
	\label{eq:q_high}
	\begin{align}
	& q(t) = \frac{a_0}{0!} + \frac{a_1}{1!} \tau 
	+ \ldots + \frac{a_{M+1}}{(M+1)!} \tau^{M+1}             \label{eq:q_high_a} \\
	& \dot{q}(t) = \frac{a_1}{0!} + \frac{a_2}{1!} \tau 
	+ \ldots + \frac{a_{M+1}}{M!} \tau^{M}                   \label{eq:q_high_b} \\
	& \vdotswithin{=} \nonumber \\
	& q^{(M)}(t) = a_M + a_{M+1}\tau,                 \label{eq:q_high_c}
\end{align}
\end{subequations}
or, more compactly as
\begin{equation}
    \label{eq:qjtau}	
    q^{(j)}(t) =  \sum_{i = j}^{M+1} \frac{a_i}{(i-j)!} \tau^{i-j},
\end{equation}
for $j=0,\ldots,M$. We then can determine $a_M$ and $a_{M+1}$ by imposing the two collocation constraints
\begin{align}
	& q^{(M)}(t_k) = g_k,\\
	& q^{(M)}(t_{k+1}) = g_{k+1},
\end{align}
where $g_k = g(\q_k,\dq_k,...,\q^{(M-1)}_k,\u_k,t_k)$. With simple calculations we find that
\begin{subequations}
	\label{eq:c_M_Mp1}
	\begin{align}
		& a_M = g_k,\\
		& a_{M+1} = \tfrac{1}{h}(g_{k+1}-g_k).
	\end{align}	
\end{subequations}
The remaining coefficients $a_0,\ldots,a_{M-1}$ are determined by imposing the initial value constraints
\begin{equation}
	q^{(j)}(t_k) = q^{(j)}_k
	\label{eq:ivc}
\end{equation} 
for $j=0,\ldots,M-1$. Using \eqref{eq:q_high} we see that the left hand side of \eqref{eq:ivc} is $a_j$, so we readily obtain
\begin{equation}
	\label{eq:cj}
	a_j = q^{(j)}_k
\end{equation} 
for $j=0,\ldots,M-1$. Finally, by evaluating \eqref{eq:qjtau} for $\tau = h$
we find that the generalized versions of the \TZ{2} formulas in \eqref{eq:TZtrans} are given by
\begin{equation}
	\label{eq:tzgeneral}
	q^{(j)}_{k+1} = \sum_{i = j}^{M+1} \frac{a_i}{(i-j)!} h^{i-j},
\end{equation}
for $j=0,\ldots,M-1$. 

One can check that, by particularizing \eqref{eq:tzgeneral} for $M=1$ and $M=2$, we obtain the equations of the trapezoidal method for first and second order systems given in \eqref{eq:tzfinval} and \eqref{eq:TZtrans}, respectively.



\newcommand{\EqsTZone}{
\begin{tabular}{l}
	${x}_{k+1} = {x_k} + \tfrac{h}{2}(f_{k+1} + f_k)$
\end{tabular}
}

\newcommand{\EqsTZtwo}{
\begin{tabular}{l}
	$\dot{q}_{k+1} = \dot{q}_k + \tfrac{h}{2}(g_{k+1}+g_k)$  
	\\ [.5em]
	${q}_{k+1} = {q_k} + \dot{q}_k h+\tfrac{h^2}{6}(g_{k+1}+ 2g_k)$
\end{tabular}
}

\newcommand{\EqsTZM}{
	\begin{tabular}{l}
	$q_{k+1}^{(M-1)} = q_k^{(M-1)} + \tfrac{h}{2}\left(g_{k+1}+g_k\right)$ 
	\\ [.5em] 
	$q_{k+1}^{(M-2)} = q_k^{(M-2)} + q_k^{(M-1)}h + \tfrac{h^2}{6}(g_{k+1}+2g_k)$ 
	\\ 
	\qquad \vdots \\
	$q^{(M-l)}_{k+1} = \left( \sum\limits_{i=0}^{l-1}\tfrac{h^i}{i!}q_k^{(i+M-l)} \right) 
	+ \tfrac{h^l}{(l+1)!} \left( l \, g_k + g_{k+1} \right)$ 
	\\ [.5em]
	\end{tabular}
}

\newcommand{\EqsHSone}{
\begin{tabular}{l}
	$x_{k+1} = x_k + \tfrac{h}{6}(f_k + 4f_c + f_{k+1})$  
	\\ [.5em]
	$x_c = \tfrac{1}{2}(x_k + x_{k+1}) + \tfrac{h}{8}(f_k - f_{k+1})$  
	\\ [.5em]
\end{tabular}
}

\newcommand{\EqsHStwo}{
\begin{tabular}{l}
	$\dot{q}_{k+1} =  \dot{q}_k + \tfrac{h}{6}( g_k+ 4g_c + g_{k+1})$ 
    \\ [.2em]
    $\dot{q}_c = \tfrac{1}{2}(\dot{q}_k+\dot{q}_{k+1})+\frac{h}{8}(g_k-g_{k+1})$ 
    \\ [1.5em]
	${q}_{k+1} = {q_k} + \dot{q}_k h+\tfrac{h^2}{6}(g_{k}+ 2g_c)$ 
	\\ [.2em]
	$q_c = q_k +\tfrac{h}{32}(13 \dot{q}_k+3 \dot{q}_{k+1})+ \tfrac{h^2}{192}(11 g_k-5 g_{k+1})$ 
	\\ [.5em]
\end{tabular}
}

\newcommand{\EqsHSM}{
\begin{tabular}{l}
	$q_{k+1}^{(M-1)} = q_k^{(M-1)} + \tfrac{h}{6}(g_k+4g_c+g_{k+1})$ 
	\\ [.6em]
	$q_c^{(M-1)} = \tfrac{1}{2} \left( q_k^{(M-1)}+q_{k+1}^{(M-1)} \right) 
	+ \frac{h}{8}(g_k-g_{k+1})$   
	\\ [2em]
	$q_{k+1}^{(M-2)} = q_k^{(M-2)} + q_k^{(M-1)} h + \tfrac{h^2}{6} (g_k+2g_c) $
    \\ [.6em]
	$q_c^{(M-2)} = q_k^{(M-2)} +\tfrac{h}{32} \left( 13 q_k^{(M-1)} + 3 q_{k+1}^{(M-1)} \right) + 
	\tfrac{h^2}{192}(11 g_k-5 g_{k+1})$  
    \\ [.8em]
	\qquad \vdots 
    \\ [.8em]
	$q_{k+1}^{(M-l)} = \left( \sum\limits_{i=0}^{l-1}\frac{h^i}{i!}q_k^{(i+M-l)} \right) 
	+ \tfrac{h^l \left(l^2g_k +4 l g_c + (2-l)g_{k+1} \right) }{(l+2)!}$
    \\ [1.2em] 
	$q^{(M-l)}_{c} = \left ( \sum\limits_{i=0}^{l-2}\frac{h^i}{2^i \, i!} q_k^{(i+M-l)} \right) 
	+ 
	\tfrac{h^{l-1} \left( 3 q_{k+1}^{(M-1)} + (2l^2+4l-3) q_k^{(M-1)} \right)}{2^l \, l! \, (l+2)}
	+ 
	\tfrac{h^l \left( (2l^2+2l-1) g_k - (2l+1) g_{k+1} \right) }{2^{l+1} \,(l+2)!}$ \\  [1.5em]
\end{tabular}
}


\begin{table*}[p]
	\small
	\begingroup
	\setlength{\tabcolsep}{7pt}
	\renewcommand{\arraystretch}{1.4} 
	\begin{tabular}{ll@{}}
		\toprule
		\textbf{Method} 	& 
		\hspace{1.5mm} \textbf{Collocation equations}
		\\ 
		\midrule
		\TZ{1}			& 
		\EqsTZone 		     					
		\\ 
		\midrule
		\TZ{2}			& 
		\EqsTZtwo			
		\\ 
		\midrule
		\TZ{M} 			& 
		\EqsTZM				
		\\ 
		\midrule
		\HS{1}		& 
		\EqsHSone    							
		\\ 
		\midrule
		\HS{2} 			& 
		\EqsHStwo				
		\\
		\midrule
        \HS{M} 			& 
		\EqsHSM
		\\			
		\bottomrule
	\end{tabular}
	\endgroup
	\vspace{1mm}
	\caption{Collocation equations for all methods of the trapezoidal and Hermite-Simpson families. For the \TZ{M} and \HS{M} methods, we provide the general equation of \scalebox{0.75}{$q_{k+1}^{(M-l)}$} (where $l$ is meant to run up to $M$) but also the particular instances of this equation for $l=1,2$. The equation of \scalebox{0.75}{$q_c^{(M-l)}$}, and its instances for $l=1,2$, are also provided in the \HS{M} method (where, again, $l=1,\ldots,M$). This arrangement allows us to realize that, within each family, the equations for the same $l$ coincide for all orders.}
    \label{table:summary}
\end{table*}

\begin{table*}[t!]
	\begingroup
	\begin{center}
		\setlength{\tabcolsep}{24pt}
		\renewcommand{\arraystretch}{1.2} 
		\begin{tabular}{@{}cccc@{}}
			\toprule
			Family & $n_v$ & $n_e$ & $n_\text{DOF}$ \\ 
			\midrule
			Trapezoidal & $(N+1) \, (n_x+n_u)$ & $n_x \, N -  n_b$ & $n_x + (N+1) \, n_u - n_b$ 
			\\
			Hermite-Simpson & $(2N+1) \, (n_x+n_u)$ & $2n_x \, N - n_b$ & $n_x + (2N+1) \, n_u - n_b$
			\\ 
			\bottomrule
		\end{tabular}
	\end{center}
	\vspace{1mm}
	\caption{Number of variables ($n_v$), equations ($n_e$), and degrees of freedom ($n_\text{DOF}$) in the two families of methods.\label{table:sizes}}
	\endgroup
\end{table*}

\subsection{The generalized Hermite-Simpson method}
\label{subsec:hs_high}

An analogous route can be followed to obtain a Hermite-Simpson method for $M$th order systems. In this case, $q^{(M)}(t)$ must be quadratic in order to determine its coefficients by imposing the collocation constraints at $t_k$, $t_{k+1}$, and $t_c = t_k + h/2$. This means that $q(t)$ must be of degree $M+2$ now, so $q(t)$, and its derivatives, will take the form
\begin{equation}
	\label{eq:qjtauHS}	
	q^{(j)}(t) =  \sum_{i = j}^{M+2} \frac{a_i}{(i-j)!} \tau^{i-j}
\end{equation}
for $j=0,\ldots,M$. The last equation in \eqref{eq:qjtauHS} is
\begin{equation} 
	\label{eq:q_high_hsc_last}
	q^{(M)}(t) = a_M + a_{M+1}\tau + \frac{a_{M+2}}{2}\tau^2,
\end{equation}
and its coefficients $a_M$, $a_{M+1}$, and $a_{M+2}$ can be determined by imposing
\begin{align}
    & q^{(M)}(t_k) = g_k,\\
    & q^{(M)}(t_c) = g_c,\\
    & q^{(M)}(t_{k+1}) = g_{k+1},
\end{align}
where
\begin{subequations}
\label{eq:gc_qc}
\begin{align}
	& g_c = g(\q_c,\dq_c,...,\q^{(M-1)}_c,\u_c,t_c),
	\label{eq:gc_qca} \\
	& \q^{(j)}_c = \q^{(j)}(t_c), \hspace{3mm} j=0,\ldots,M-1.
	\label{eq:gc_qcb}
\end{align}	
\end{subequations}
After simple calculations we find that
\begin{subequations}
	\label{eq:last_coefs_HS}
	\begin{align}
		& a_M = g_k, \label{eq:last_coefs_HSa}\\
		& a_{M+1} = -\tfrac{1}{h}\left(3g_k - 4g_c + g_{k+1}\right), \label{eq:last_coefs_HSb}\\
		& a_{M+2} = \tfrac{4}{h^2}(g_k-2g_c+g_{k+1}). \label{eq:last_coefs_HSc}
	\end{align}	
\end{subequations}
As in the trapezoidal method, the remaining coefficients are determined by the initial value constraints, and we have
\begin{equation}
	\label{eq:remaining_coefs_HS}
	a_j = q^{(j)}_k,
\end{equation} 
for $j=0,\ldots,M-1$.
The generalized versions of Eqs.~\eqref{eq:HS2cont} can then be obtained by evaluating the expressions up to order $M-1$ in \eqref{eq:qjtauHS} for $\tau = h$, and using $q^{(j)}(t_{k+1}) = q^{(j)}_{k+1}$. This yields
\begin{equation}
	\label{eq:HSgeneral_a}
	q^{(j)}_{k+1} = \sum_{i = j}^{M+2} \frac{a_i}{(i-j)!} h^{i-j}
\end{equation}
for $j=0,\ldots,M-1$.

As it happens in the Hermite-Simpson method for 2nd order systems, $g_c$ in \eqref{eq:last_coefs_HS} requires the midpoint values $\q_c, \dq_c, \ldots, \q_c^{(M-1)}$, but these are easily obtained by evaluating \eqref{eq:qjtauHS} for $\tau = h/2$, which results in
\begin{equation}
	\label{eq:HSgeneral_b}
	q^{(j)}_c = \sum_{i = j}^{M+2} \frac{a_i}{(i-j)!} \left(\frac{h}{2}\right)^{i-j}
\end{equation}
for $j=0,\ldots,M-1$. 

The terms $a_{M+1}$ and $a_{M+2}$ in \eqref{eq:HSgeneral_b} involve $g_c$ and thus the midpoint coordinate $q_c$ and its derivatives. However, we can remove the dependence of $q^{(j)}_c$ on $g_c$ by using the last equation in \eqref{eq:HSgeneral_a}, which is
\begin{equation}\label{eq:HSgeneral_a_last}
	q^{(M-1)}_{k+1} = q^{(M-1)}_k + \frac{h}{6}(g_k + 4 g_c +g_{k+1}).
\end{equation}
By isolating $g_c$ from this equation we have
\begin{equation}\label{eq:gc_general}
	g_c = \frac{g_{k+1} - g_k}{4} + \frac{3 q_{k+1}^{(M-1)} - 3 q_{k}^{(M-1)}}{2h},
\end{equation}
which we can substitute in the expressions of $a_{M+1}$ and $a_{M+2}$ involved in \eqref{eq:HSgeneral_b}. With these substitutions applied, \eqref{eq:HSgeneral_a} and \eqref{eq:HSgeneral_b} form a separated form of the Hermite-Simpson method for $M$th order systems. The condensed form is finally achieved by substituting the new version of \eqref{eq:HSgeneral_b} in the expressions of $a_{M+1}$ and $a_{M+2}$ in 
\eqref{eq:HSgeneral_a}. 

Again, one can verify that, for $M=1$ and \mbox{$M=2$}, Eqs.~\eqref{eq:HSgeneral_a} and \eqref{eq:HSgeneral_b} yield the Hermite-Simpson formulas for first and second order systems given in \eqref{eq:HSsep1}, and in \eqref{eq:HS2cont} and \eqref{HS2c_alt}, respectively.

\section{Comparison of the methods}
\label{sec:summary}

Table~\ref{table:summary} summarizes the equations for all methods of the trapezoidal and Hermite-Simpson families. For short, we refer to the methods in each family by \TZ{} and \HS{}, followed by a number that indicates the order assumed for the system dynamics. For the general \TZ{M} and \HS{M} methods, the table provides the equations for \scalebox{0.85}{$q_{k+1}^{(M-l)}$}, as well as \scalebox{0.85}{$q_{k+1}^{(M-l)}$} and \scalebox{0.85}{$q_c^{(M-l)}$}, respectively, where $l$ runs from $1$ to $M$ in all cases. We also specialize these equations for $l=1,2$, so the reader can realize that, within each family, the equations for a same value of $l$ coincide for all orders. 

In the table, the equations for the Hermite-Simpson methods are given in their separated form, and in the \HS{M} method we show those that result from applying the manipulations described in Section~\ref{subsec:hs_high}. 

\subsection{Problem size}
\label{subsec:problem_sizes}

It is not difficult to see that, for all methods in a same family, the number of variables ($n_v$), equations ($n_e$), and degrees of freedom ($n_\text{DOF}$) is the same in the resulting transcriptions of Problem \eqref{eq:OCP}. If $n_x$ and $n_u$ are the dimensions of $\x$ and $\u$, and $n_b$ is the number of boundary constraints in Eq.~\eqref{eq:OCP_boundary}, we obtain the values in Table~\ref{table:sizes}.
Note that for an $M$-th order ODE, the state $\x$ includes the configuration vector $\q$ and its derivatives, so that $n_x=M n_q$, where $n_q$ is the dimension of $\q$.
The improved formulas, as compared to those of the \TZ{1} and \HS{1} methods, neither increase the problem size, nor reduce the freedom to find the optimal solution. Moreover, since the dynamic function must be evaluated at each collocation point, the number of  
evaluations is the same in all methods of a same family, so the new methods should not increase the cost of each iteration when solving the transcribed NLP problem. This point is also supported by the computational experiments that we present in Section~\ref{sec:tests}.

\subsection{Accuracy of the approximations}
\label{subsec:accuracy}

While the new methods introduced in this paper are explicitly designed to preserve the consistency between the configuration trajectory and its derivatives, a further question is how the application of these methods may affect the accuracy of the solution approximations and its rate of convergence as $h \rightarrow 0$. To answer this question, we draw upon the concept of order of accuracy \citep{betts2010practical}, or simply order \citep{hairer2002geometric} of a collocation method, which, in turn, relies on the definition of local error of an approximation.

The local error $\epsilon_k$ of a collocation method at interval $k$ is defined as the difference between the computed value $q_{k+1}$ and the value for $t=t_{k+1}$ of the exact solution of the ODE, $\hat{q}(t)$, that passes through the computed point $q_k$. If a collocation method approximates the solution with polynomials of degree $d$, we have for interval $k$:
\begin{equation}
    q(t) = q_k + a_1 \tau + \frac{a_2}{2} \tau^2 + \dots + \frac{a_d}{d!} \tau^d, \nonumber
\end{equation}
and the computed value $q_{k+1}$ is obtained by setting $t=t_k+h$, which means setting $\tau=h$:
\begin{equation}
\label{eq:taylqh}
    q_{k+1} = q_k + a_1 h + \frac{a_2}{2}  h^2 + \dots + \frac{a_d}{d!} h^d.
\end{equation}
On the other hand, the Taylor expansion of the exact solution $\hat{q}(t)$  that passes through the computed point $q_k$ is
\begin{equation}
\nonumber
	\label{eq:taylx}
	\begin{split}
	\hat{q}(t_k+t) = q_k + \dot{\hat{q}}(t_k) \; \tau 
	+ 
	\frac{\ddot{\hat{q}}(t_k)}{2} \tau^2 + \dots
	+ \\
	+ \frac{\hat{q}^{(d)}(t_k)}{d!} \tau^d
	+
	\bigO(\tau^{d+1}),
	\end{split}
\end{equation}
and evaluating for $\tau=h$ we have:
\begin{equation}
	\label{eq:taylxh}
	\begin{split}
	\hat{q}(t_k+h) = q_k + \dot{\hat{q}}(t_k) \; h
	+ 
	\frac{\ddot{\hat{q}}(t_k)}{2} h^2 + \dots
	+ \\
	+ \frac{\hat{q}^{(d)}(t_k)}{d!} h^d
	+
	\bigO(h^{d+1}),
	\end{split}
\end{equation}
thus, the local error $\epsilon_k$ is given by the difference of the two Taylor expansions (\ref{eq:taylqh}) and (\ref{eq:taylxh}):
\begin{equation}
	\label{eq:tayle}
	\begin{split}
        \epsilon_k = 
        \left(a_1 - \dot{\hat{q}}(t_k)\right)  h 
        +
        \frac{a_2 - \ddot{\hat{q}}(t_k)}{2} h^2 
        + \dots + 
        \\
        + \frac{a_d - \hat{q}^{(d)}(t_k)}{d!} h^d
        + \bigO(h^{d+1}).
	\end{split}
\end{equation}

A collocation method is said to have order of accuracy $p$ if the sum of the first $p$ terms of (\ref{eq:tayle}) is zero. Note that this does not imply that each term vanishes by itself: when $h$ takes a specific numerical value, different non-null terms of the sum may add to zero. In the hypothetical case that the exact solution $\hat{q}(t_k + t)$ was a polynomial of degree $p$, a method of order $p$ would have no local error. For this reason, the order of accuracy is also called the degree of exactness~\citep{Dalquist2008}, and an equivalent definition for it is that a collocation method has order of accuracy $p$ if it is exact for all polynomials of degree $\leq p$.

In the limit, when $h \rightarrow 0$, the error of the approximation will converge to zero. The rate of this convergence is an important property of a method, and is directly given by its order. If a method has order $p$, the lower power of $h$ appearing in $\epsilon_k$ is $p+1$, so that, when $h \rightarrow 0$, the local error decreases as $h^{p+1}$:
\vspace*{-2mm}
\begin{equation}
    \nonumber
    \epsilon_k = \bigO(h^{p+1}).
\end{equation}

\vspace*{-2mm}
In all collocation methods discussed here, the interpolating polynomial used in each interval has degree $d=M+s-1$, where $M$ is the order of the ODE and $s$ is the number of collocation points of each interval. This value $d$ ensures the unique determination of the $d+1$ polynomial coefficients given the $s$ collocation constrains and the $M$ initial conditions. In the event that the exact solution happens to be a polynomial $\hat{q}(t)$ of degree $d$, it must necessarily coincide with the interpolatory polynomial, and $q_{k+1}$ will coincide with the exact value $\hat{q}(t_k+h)$. This shows that the order of accuracy of any method is at least $p = d = M+s-1$, so the orders of \TZ{1} and \HS{1} are at least 2 and 3, respectively, while the orders of \TZ{2} and \HS{2} are at least 3 and 4. However, these lower bounds can be surpassed in some cases. For example, the \HS{1} method is known to have order 4~\citep{hairer2002geometric}, while its corresponding lower bound is 3. This is because it takes advantage of a special property of a family of polynomials of fourth degree. It can be proved that any fourth degree polynomial satisfying 
\begin{subequations}
\label{eq:cond_4th_deg}
\begin{align}
&\hat{q}(t_k)={q}_k,
\\
&\dot{\hat{q}}(t_k)= \dot{{q}}_k,
\\
&\dot{\hat{q}}(t_k+h/2)= \dot{{q}}_c,
\\
&\dot{\hat{q}}(t_k+h)= \dot{{q}}_{k+1}
\end{align}
\end{subequations}
takes always the same value $\hat{q}(t_k+h)={q}_{k+1}$. Since the only third degree polynomial satisfying these same conditions is a particular case of this family, it satisfies $q(t_k+h)={q}_{k+1}$, so its order of accuracy is 4.

\begin{figure*}[t!]
	\begin{center}
		\includegraphics[width=1\linewidth]{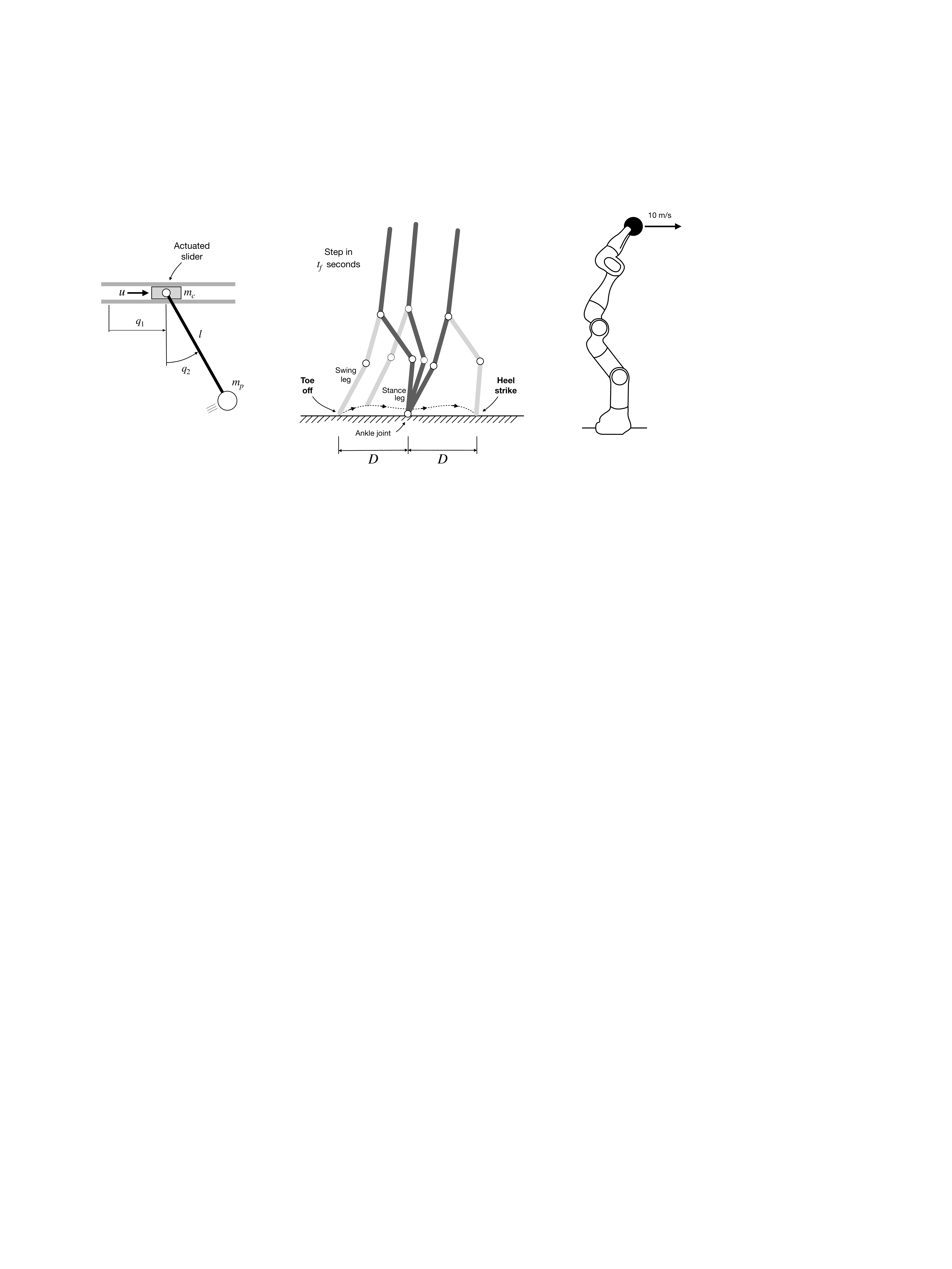}		
	\end{center}
	\caption{\label{fig:testcases}Benchmark problems. Left: A cart-pole system that has to perform a swing-up motion. Center: a walking biped whose periodic gait must be optimized (the three snapshots illustrate the motion that occurs between the \emph{toe off} and \emph{heel strike} events defining a period of the gait). Right: A 7R Panda robot that has to pick a ball at the shown configuration, and throw it from the same configuration at $10$m/s horizontally.}
\end{figure*}

Even if the \HS{2} method does not benefit from a similar property, it is granted that its order is at least as large as that of \HS{1}, i.e., 4. 

So, we can say that the order of accuracy of the presented methods for second order systems is equal or higher than that of the corresponding methods for first order systems. In general, for the same number of collocation points, a method for $M$th order systems has this lower bound $M-1$ units higher than the corresponding method for first order systems.

\section{Test cases}
\label{sec:tests}

\begin{figure*}[p]
	\begin{center}
		\includegraphics[width=0.48\linewidth]{
            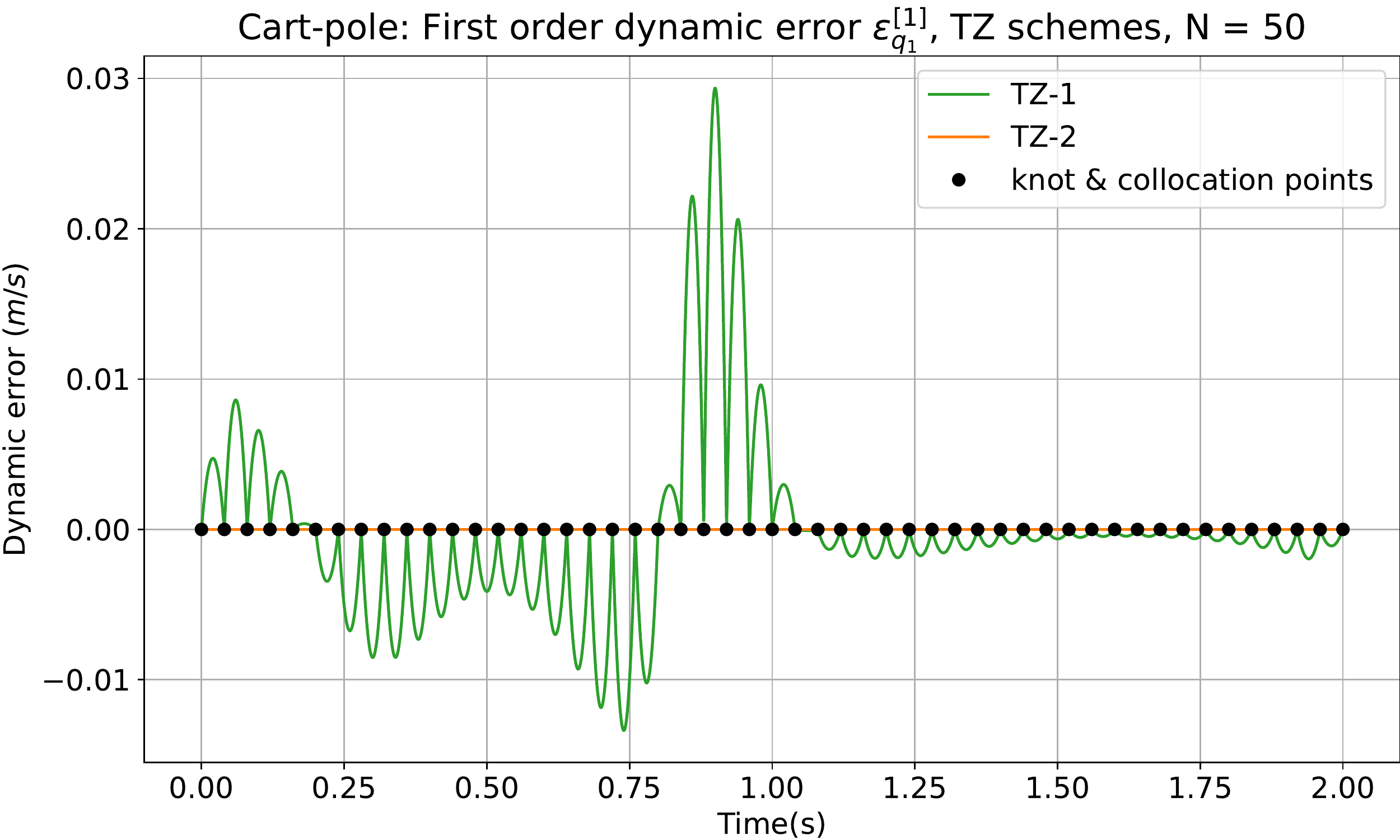}
		\hspace{2mm}	    
		\includegraphics[width=0.48\linewidth]{
			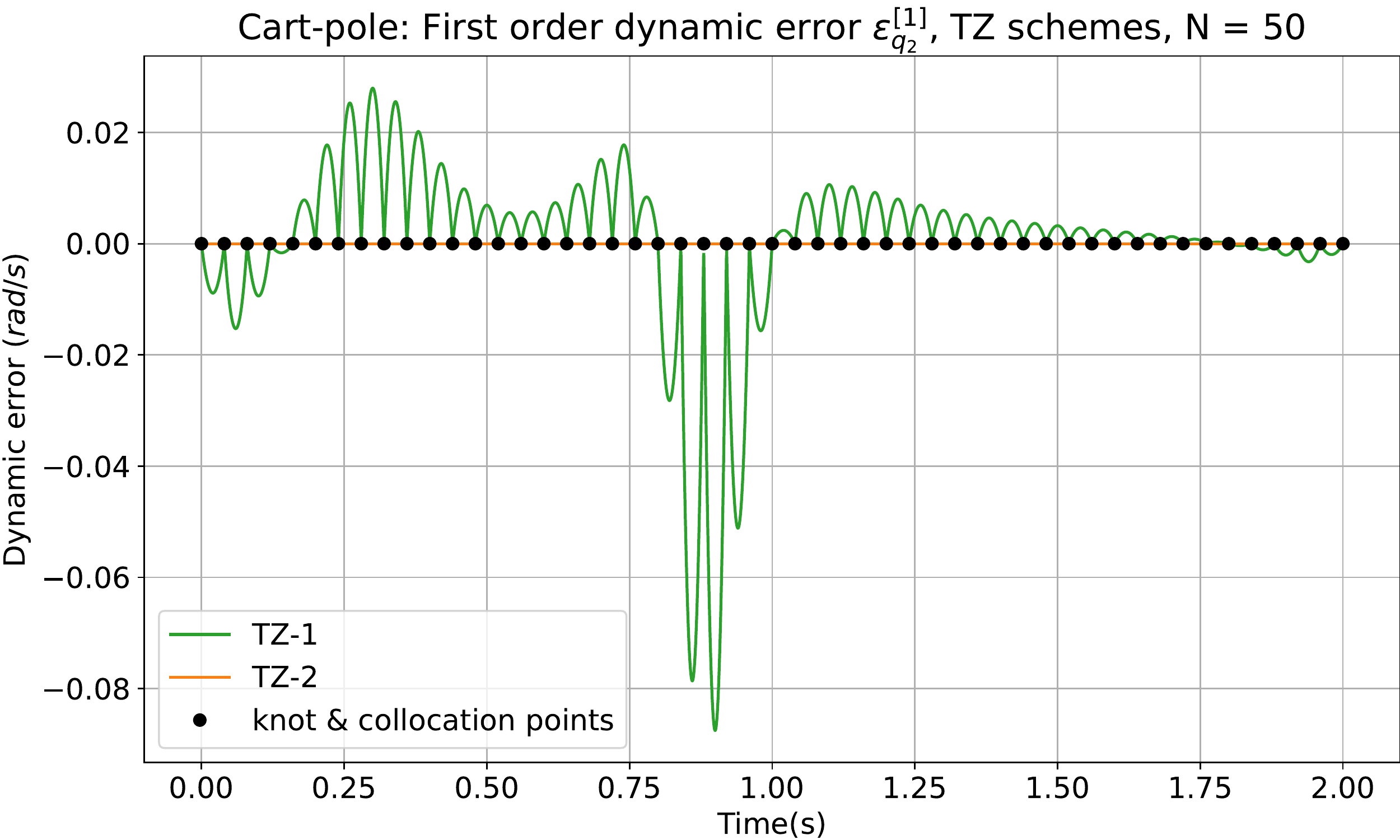}
		\vspace{2mm}
		\includegraphics[width=0.48\linewidth]{
			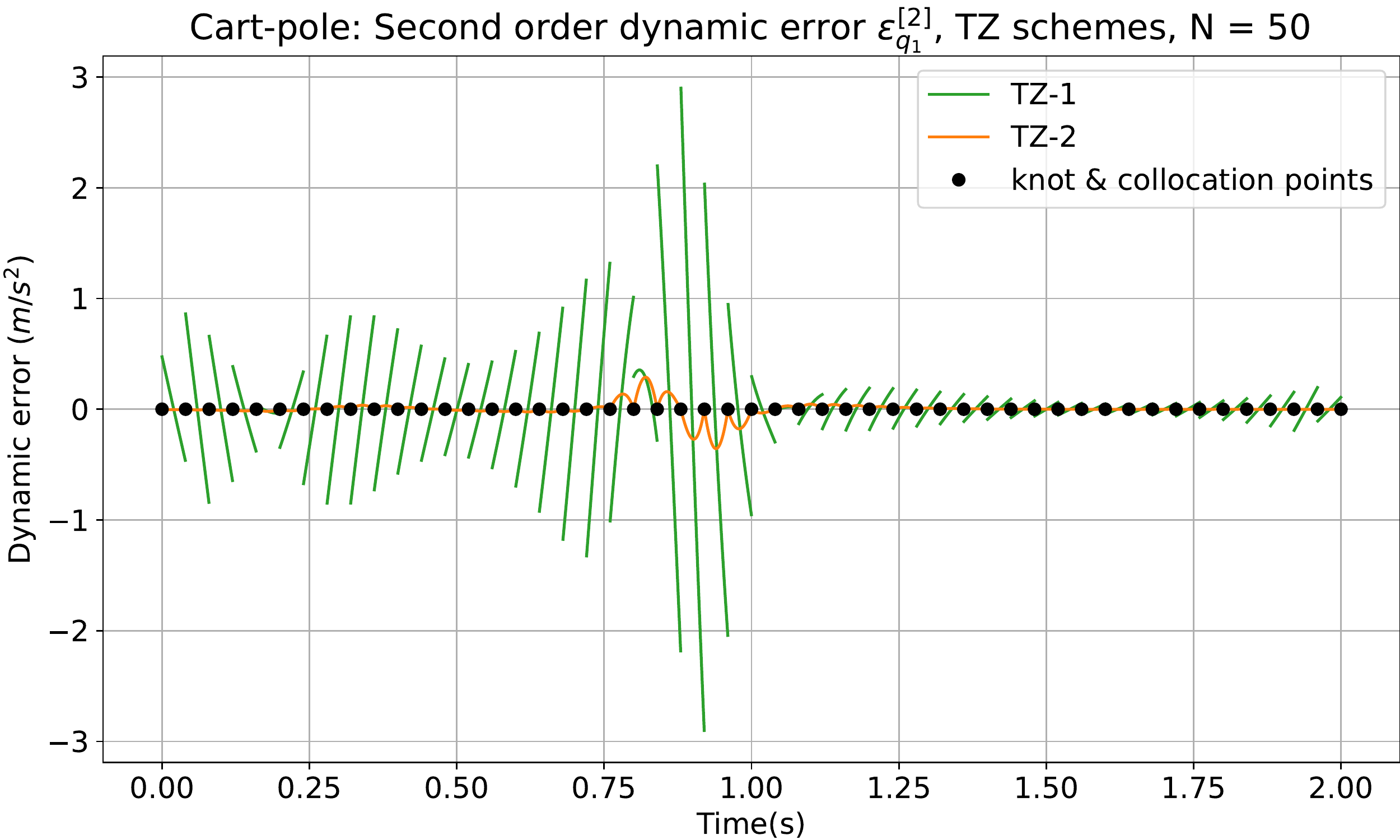}
		\hspace{2mm}	    
		\includegraphics[width=0.48\linewidth]{
			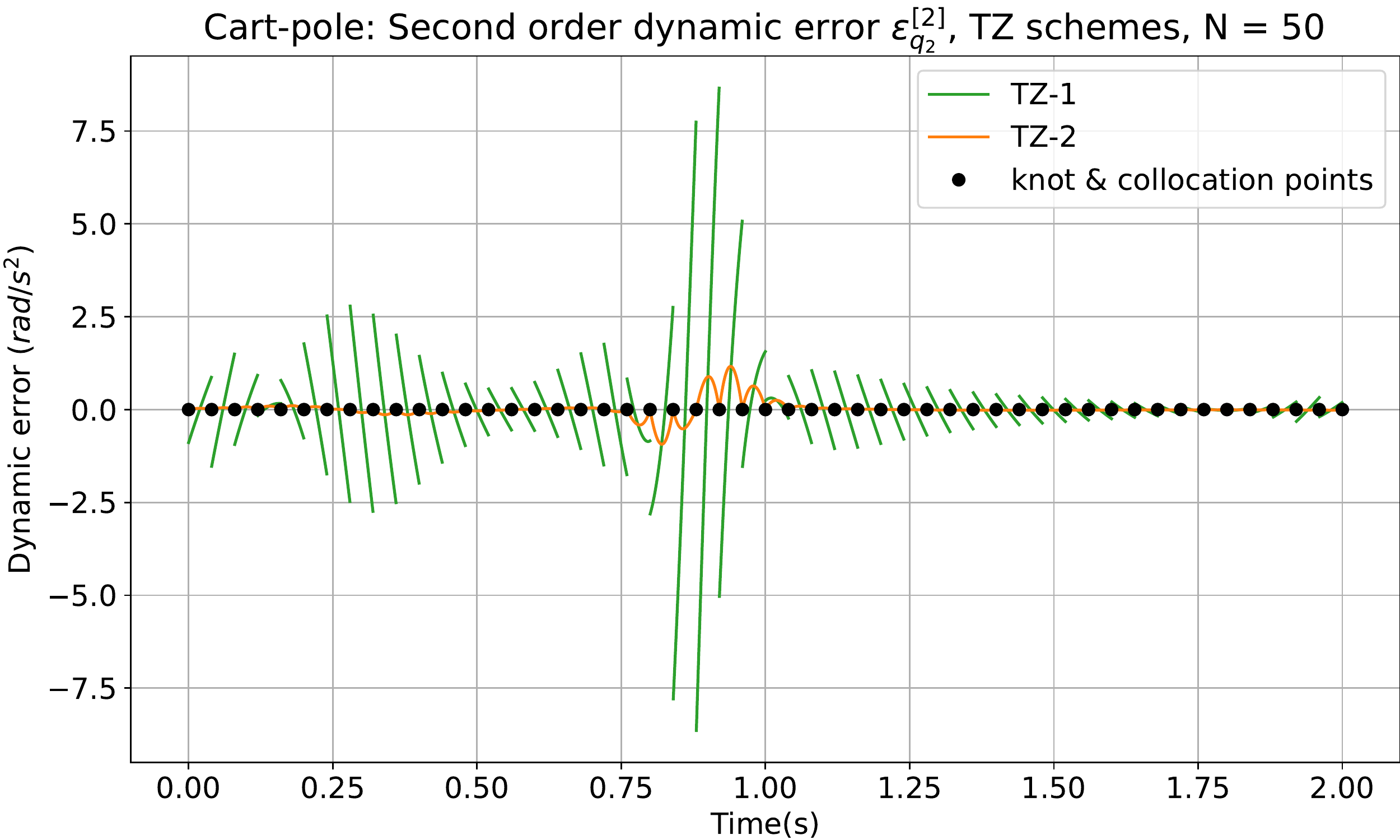}
		\vspace{2mm}
		\includegraphics[width=0.48\linewidth]{
			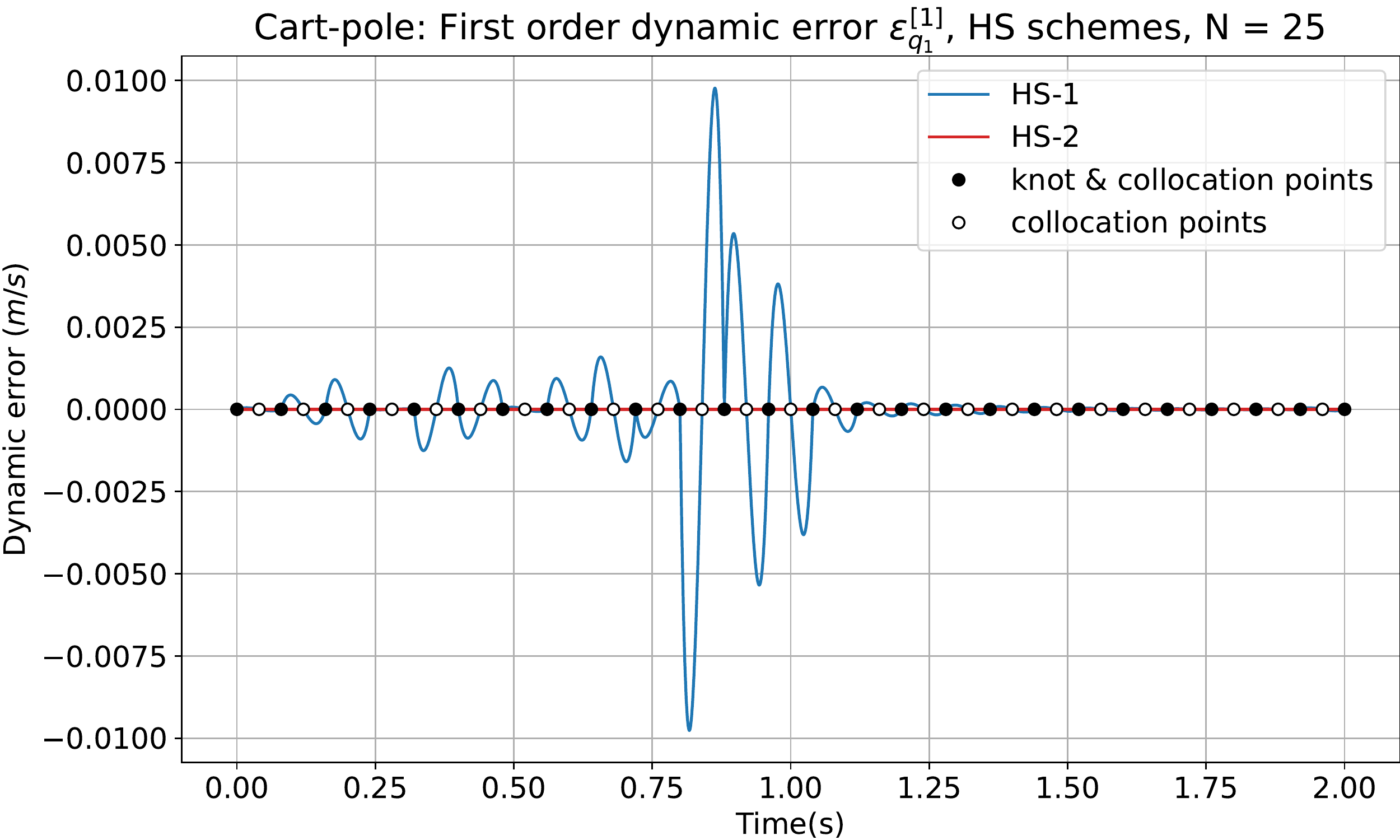}
		\hspace{2mm}	    
		\includegraphics[width=0.48\linewidth]{
			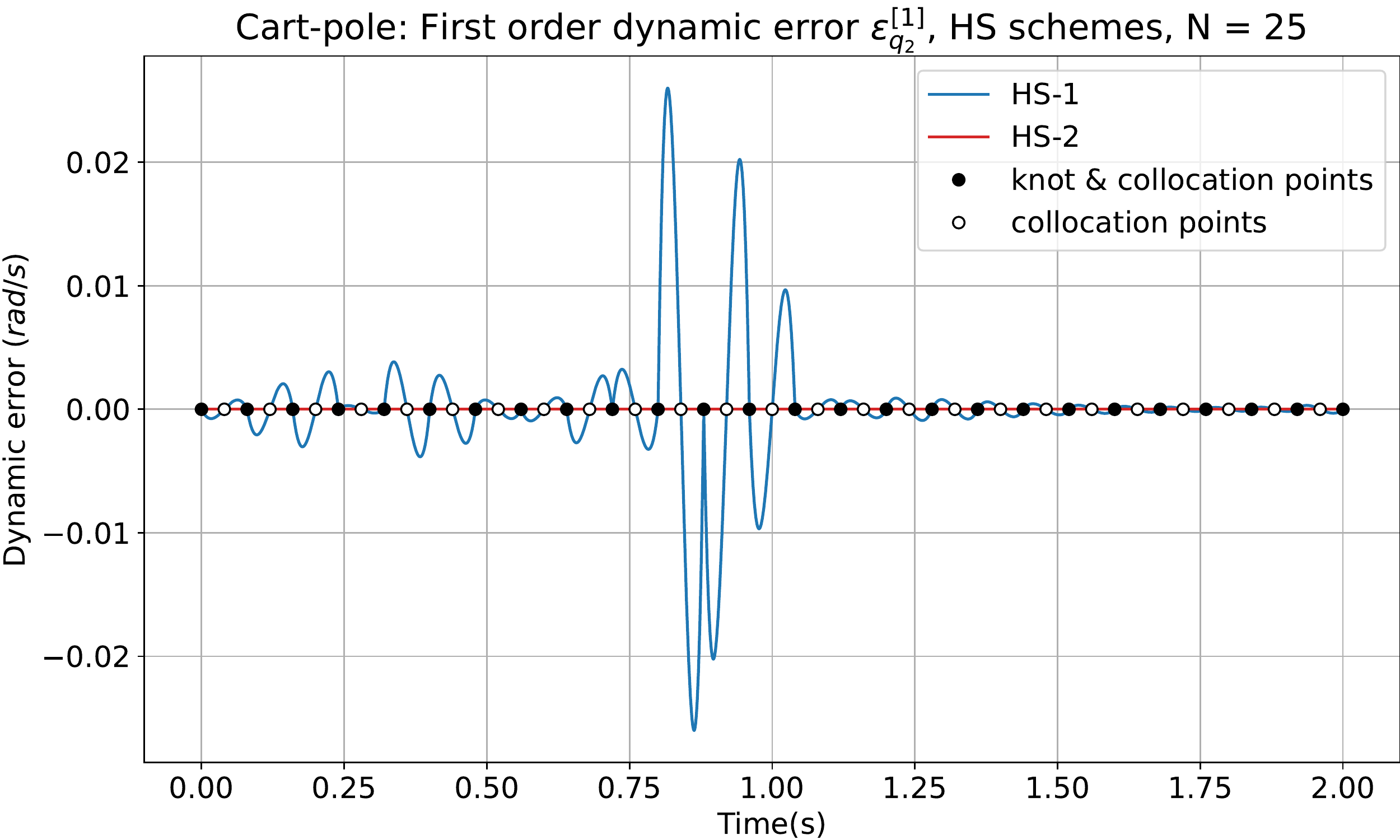}
		\vspace{2mm}
		\includegraphics[width=0.48\linewidth]{
			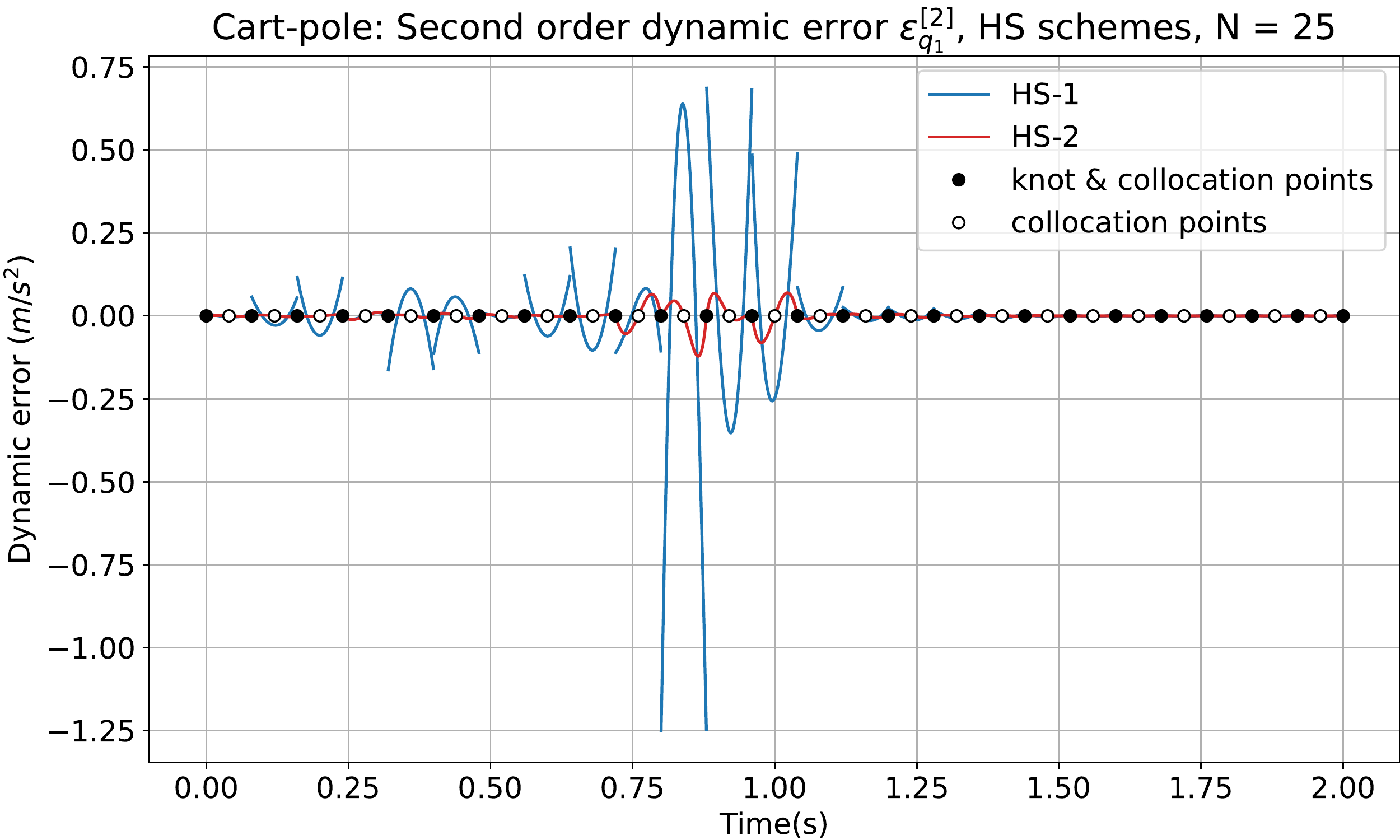}
		\hspace{2mm}	    
		\includegraphics[width=0.48\linewidth]{
			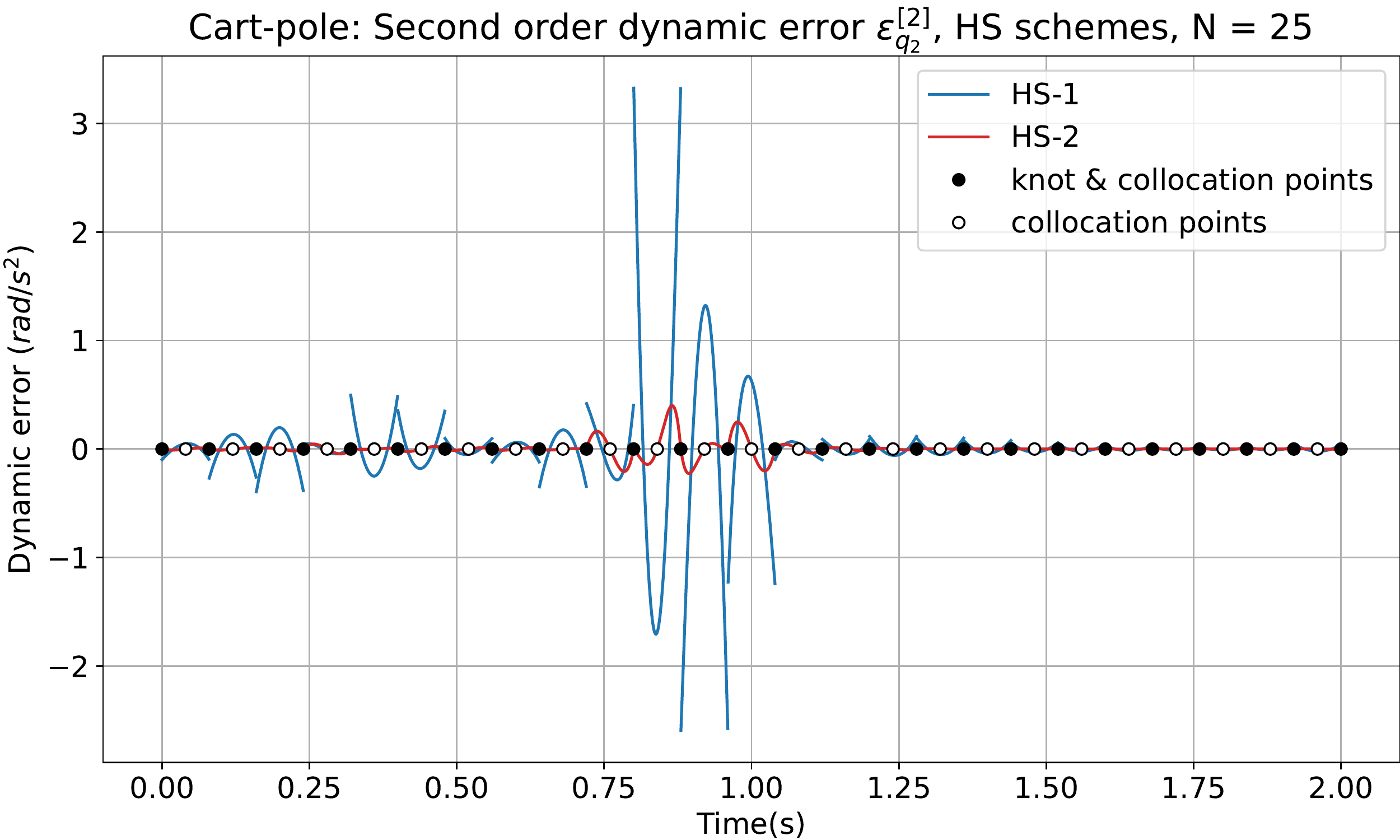}
	\end{center}
	\caption{Cart-pole problem: Plots of the first and second order dynamic errors $\varepsilon^{[1]}_{q_i}(t)$ and $\varepsilon^{[2]}_{q_i}(t)$ for $q_1$ and $q_2$ (left and right columns, respectively), using the trapezoidal and Hermite-Simpson methods. To compare the results with those by \cite{kelly2017introduction}, note that the latter paper actually provides the plots of \scalebox{0.9}{$-\varepsilon^{[1]}_{q_i}(t)$} for \HS{1}.
	\label{fig:CP_graphs}}
\end{figure*}

The performance of all methods is next evaluated and compared using three trajectory optimization problems shown in Fig.~\ref{fig:testcases}. We refer to them as the cart-pole, bipedal walking, and ball throwing problems, respectively. The first two problems are solved and documented in detail by \citet{kelly2017introduction}, and thus serve to compare our results with those published in the literature. The third problem is proposed by the authors to illustrate the methods on a widely-used robot with a complex dynamics. Since analytical solutions for these problems are not available, we compare the different methods by computing the dynamic transcription errors they produce. To this end, we define the following errors relative to the dynamics constraints.

The first order dynamic error for the $q_i$ coordinate is defined as
\begin{align}
	\label{eq:1st_order_error}
	\varepsilon^{[1]}_{q_i}(t) = \dot{q}_i(t) - v_i(t).
\end{align}
In general, this error is non-null in \TZ{1} and \HS{1}, as these methods do not enforce $v_i(t)=\dot{q}_i(t)$ for all $t$. For the same coordinate, the second order dynamic error is
\begin{align}
	\label{eq:2nd_order_error}
	\varepsilon^{[2]}_{q_i}(t) = \ddot{q}_i(t) - g_i(\vr{q},\vr{\dot{q}},\vr{u}, t).
\end{align}
We found this error to be more meaningful than the $\varepsilon^{[1]}_{q_i}(t)$ error reported by \citet{kelly2017introduction}, since it reflects the  deviation from the actual system dynamics, which is expected to be minimized with the optimization process. Another error that is useful to define when all coordinates of $\vr{q}$ have the same units is the joint error
\begin{align}
	\varepsilon^{[r]}(t) = 
	\lVert\varepsilon^{[r]}_{q_1}(t)\rVert + \ldots + \lVert\varepsilon^{[r]}_{q_{n_q}}(t)\rVert,
\end{align}
for $r=1,2$. Finally, to summarize the error functions in just one number, we can compute their integrals over $[0,t_f]$:
\begin{align}
	E^{[r]}_{q_i} &= \int_{0}^{t_f} \lVert\varepsilon^{[r]}_{q_i}(t)\rVert \; dt, & r=1,2,\\
	E^{[r]} &= \int_{0}^{t_f} \varepsilon^{[r]}(t) \; dt, & r=1,2.
\end{align}


To compare the methods, we have implemented them in Python, using the toolbox CasADi \citep{andersson2019casadi} to solve the constrained optimization problems that result. CasADi provides the necessary means to formulate such problems and to compute the gradients and Hessians of the transcribed equations using automatic differentiation. These are necessary to solve the optimization problems, a task for which we rely on the interior-point solver IPOPT \citep{wachter2006ipopt} in conjunction with the linear solver MUMPS \citep{amestoy2001mumps}. The whole implementation can be downloaded 
from \url{https://github.com/AunSiro/optibot}, but the reader can also reproduce the results for the cart-pole and bipedal walking problems through interactive Jupyter notebooks online \citep{authors2022jupytercartpole,authors2022jupyterbipedal}. The execution times we report have been obtained on a single-thread implementation running on an iMac computer with an Intel i7, 8-core 10th generation processor at 3.8 GHz.

\subsection{The cart-pole swing-up problem}
\label{subsec:cart_pole}

The cart-pole system comprises a cart that travels along a horizontal track and a pendulum that hangs freely from the cart. A motor drives the cart forward and backward along the track. Starting with the pendulum hanging below the cart at rest at a given position, the goal is to reach a final configuration in a given time $t_f$, with the pendulum stabilized at a point of inverted balance and the cart staying at rest at a distance $d$ from the initial position. The cost to be minimized is
\begin{equation}  \label{J}
	\int_{0}^{t_f}u^2(t) dt,
\end{equation}
where $u$ is the force applied to the cart, and we adopt the same dynamic equations and problem parameters as in \citet{kelly2017introduction}. An animation of the solution obtained with \HS{2} and $N=25$ can be seen in
\url{https://youtu.be/M0ivg_8s-I8}.

Figure \ref{fig:CP_graphs} compares the errors $\varepsilon^{[1]}_{q_i}(t)$ and $\varepsilon^{[2]}_{q_i}(t)$ obtained by the methods for the variables $q_1$ and $q_2$ shown in Fig.~\ref{fig:testcases}. The number $N$ of intervals used in the comparison is $50$ for the trapezoidal scheme, and $25$ for the Hermite-Simpson one. This yields a fair comparison, as then the number of collocation points, variables, and degrees of freedom of the optimization are the same in all NLP problems (cf. Table \ref{table:sizes}). The plots corresponding to the first order error $\varepsilon^{[1]}_{q_i}(t)$ of Fig. \ref{fig:CP_graphs}, in the first and third row respectively, confirm that \TZ{1} and \HS{1} present a non-negligible first order error, while in \TZ{2} and \HS{2} this error is exactly zero as expected.

The plots in the second and fourth rows clearly show a discontinuity at the knot points of the second order error $\varepsilon^{[2]}_{q_i}(t)$ for \TZ{1} and \HS{1}, reflecting the discontinuity of $\ddq(t)$ at these points. In contrast, for \TZ{2} and \HS{2}, the error functions are continuous and vanish at the collocation points, evidencing that, as anticipated in Section~\ref{subsec:drawbacks}, the system dynamics is exactly satisfied at all collocation points for the new methods, but not for the conventional ones.

The figure also shows the dramatic reductions of $\varepsilon^{[2]}_{q_i}(t)$ for the new methods when compared with the corresponding \TZ{1} and \HS{1} ones. The numerical evaluation of the results appears in Table \ref{table:CPerrors}, which provides the computation times $t_c$ and the integral errors $E^{[2]}_{q_i}$ for this problem.  It can be seen that the errors $E^{[2]}_{q_i}$ are almost one order of magnitude lower for \TZ{2} and \HS{2} than for their counterparts \TZ{1} and \HS{1}, despite using a very similar computation time. It is interesting to see that the errors $E^{[2]}_{q_i}$ achieved by \TZ{2} are about a half of those of \HS{1} for the same number of collocation points. The comparison is relevant since both methods use polynomials of the same degree to approximate $q_i(t)$.

\begin{table}[tb]
	\begingroup
	\begin{center}
		\setlength{\tabcolsep}{6pt}
		\begin{tabular}{@{}ccccccc@{}}
			\toprule
			& $N$ & $t_c$ & $E^{[1]}_{q_1}$ & $E^{[1]}_{q_2}$ & $E^{[2]}_{q_1}$ & $E^{[2]}_{q_2}$ \\ [.3em]
			&    & $(s)$ & $(m)$ & $(rad)$ & $(m/s)$ & $(rad/s)$\\
			\midrule
			\TZ{1} & 50 & 0.025 & 0.0066 & 0.0167 & 0.504  & 1.281  \\
			\TZ{2} & 50 & 0.025 & 0      & 0      & 0.052  & 0.170  \\
			\HS{1} & 25 & 0.020 & 0.0014 & 0.0043 & 0.113  & 0.338  \\
			\HS{2} & 25 & 0.023 & 0      & 0      & 0.016  & 0.052  \\ \bottomrule
		\end{tabular}
	\end{center}
	\vspace{1mm}
	\caption{Performance data for the cart-pole problem.
		\label{table:CPerrors}}
	\endgroup
\end{table}

\begin{figure*}[t!]
	\begin{center}
		\includegraphics[width=0.49\linewidth]{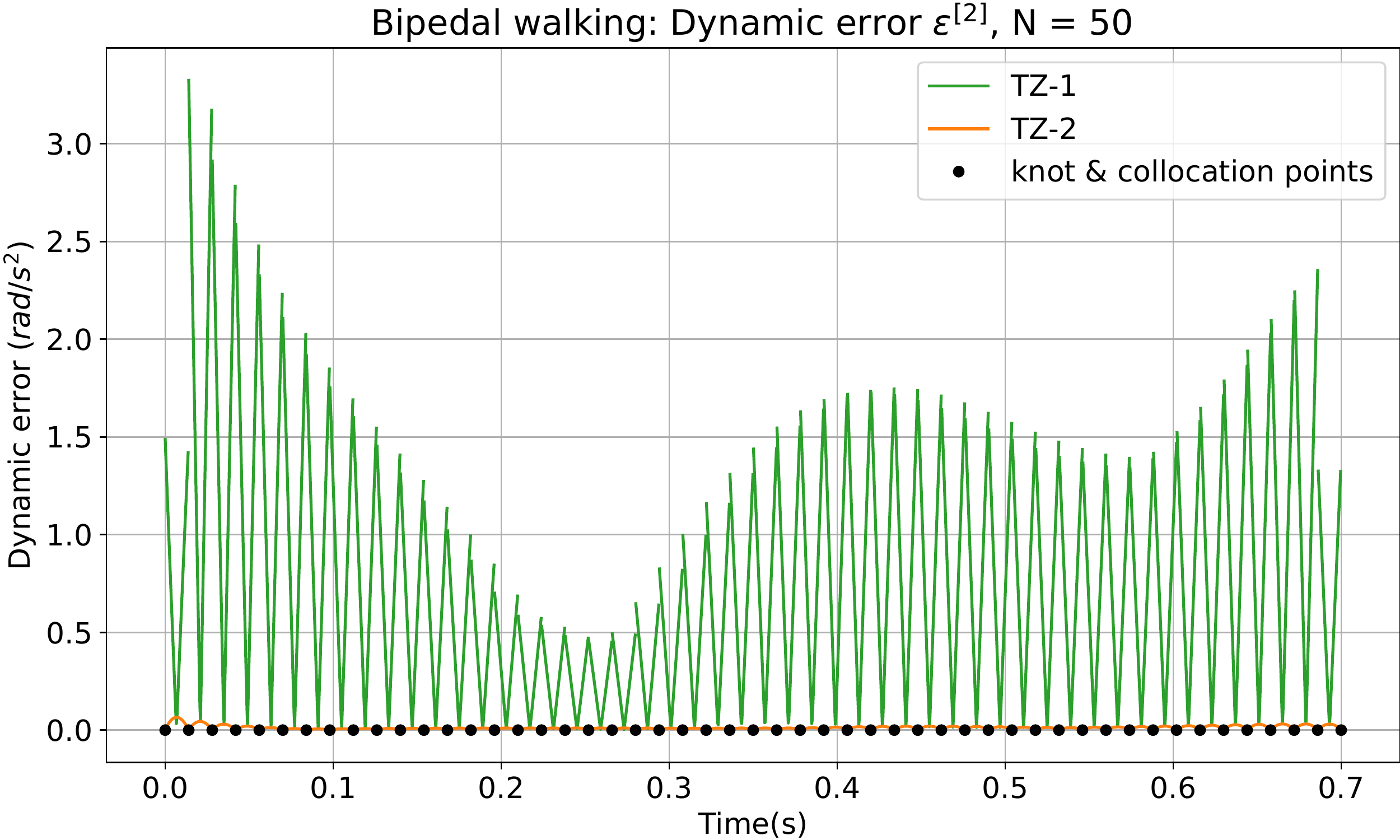}
		\includegraphics[width=0.49\linewidth]{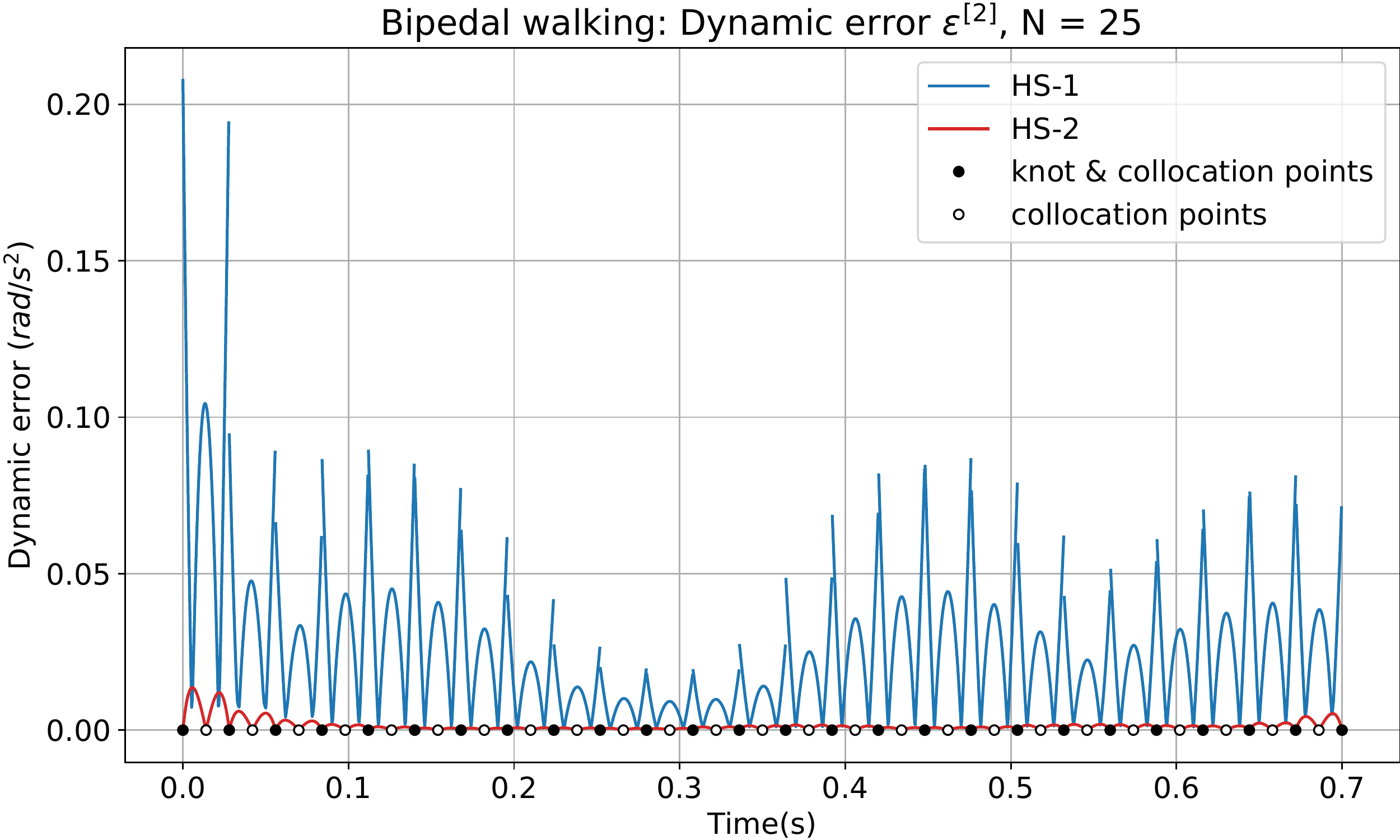}
		\vspace{2mm}
	\end{center}
	\caption{Second order dynamic errors for the bipedal walking problem.}
	\label{fig:5L_dyn_err}
\end{figure*}

\subsection{The bipedal walking problem}
\label{subsec:rabbit}

We next apply the methods to optimize a periodic gait for the planar biped robot shown in Fig.~\ref{fig:testcases}. The robot involves five links pairwise connected with revolute joints, forming two legs and a torso. All joints are powered by torque motors, with the exception of the ankle joint, which is passive. Like the cart pole system, therefore, this robot is underactuated, but it is substantially more complex. The system is commonly used as a testbed when studying bipedal walking \citep{westervelt2003hybrid,yang2009framework,park2012switching,saglam2014robust}. 

\begin{table}[t!]
	\begingroup
	\begin{center}
		\setlength{\tabcolsep}{12pt}
		\begin{tabular}{@{}ccccc@{}}
			\toprule
			Method & $N$& $t_c$ & $E^{[1]}$ & $E^{[2]}$  \\ [.3em]
			&    & $(s)$ & $(rad)$ & $(rad/s)$\\
			\midrule
			\TZ{1} & 50 & 0.122 & 0.0025  & 0.5328 \\
			\TZ{2} & 50 & 0.125 & 0       & 0.0081 \\
			\HS{1} & 25 & 0.131 & 8.2$\times10^{-5}$ & 0.0182 \\
			\HS{2} & 25 & 0.127 & 0       & 0.0011 \\ \bottomrule
		\end{tabular}
	\end{center}
    \vspace{1mm}
	\caption{Performance data for the bipedal walking problem ($E^{[2]}$ is the integral of $\varepsilon^{[2]}$ in Fig.~\ref{fig:5L_dyn_err}).
		\label{table:5Lerrors}}
	\endgroup
\vspace*{-2mm}
\end{table}

For this example we use the dynamic model given by \citet{kelly2017introduction}, which matches the one in \cite{westervelt2003hybrid} with parameters corresponding to the RABBIT prototype~\citep{chevallereau2003rabbit}. We assume the robot is left-right symmetric, so we can search for a periodic gait using a single step, as opposed to a stride, which involves two steps. This means that the state and torque trajectories will be the same on each successive step.

As in \cite{kelly2017introduction}, we define $\vr{q}$ as the vector that contains the absolute angles of all links relative to ground, while $\vr{u}$ encompasses all motor torques. Also as in \cite{kelly2017introduction}, and similarly to the cart-pole problem, our goal is to find state and action trajectories $\vr{x}(t)$ and $\vr{u}(t)$ that define an optimal gait under the cost
\vspace*{-1mm}
\begin{equation}  \label{Jbiped}
	\int_{0}^{t_f}\vr{u}(t)\trans\vr{u}(t) \; dt.
\end{equation}

\begin{figure*}[t]
	\begin{center}
		\newcounter{row}
		\newcounter{col}
		\newcounter{n}
		\newcounter{im_id}
		\forloop{row}{0}{\value{row} < 2}{
			\forloop{col}{0}{\value{col} < 5}{
				\setcounter{n}{\value{row}*5+\value{col}}
				\setcounter{im_id}{\value{n}}
				\includegraphics[width=0.2\linewidth]{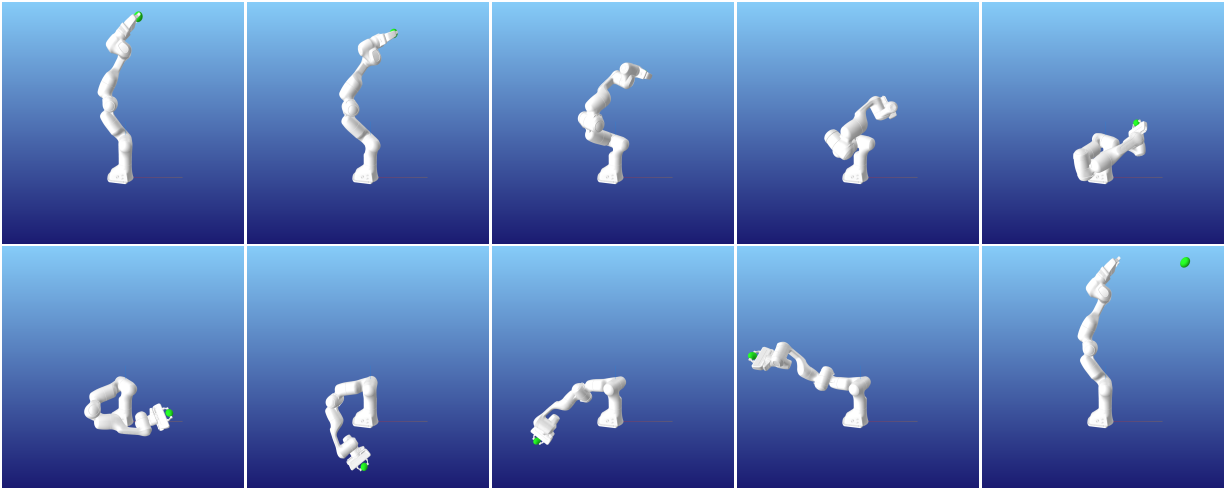}\hspace*{-1.2em}
			}
			\\
		}
	\end{center}
	\vspace{1mm}	
	\caption{Trajectory obtained for the ball throwing task. The robot, initially at rest, progressively gains momentum assisted by gravity, so as to get back to the initial configuration to throw the ball at the required speed. An animation of the trajectory can be seen in \url{https://youtu.be/NsEv6JrSN8c}. \label{fig:panda_sequence}}
\end{figure*}

\vspace*{-1mm}
Several constraints are added to ensure a feasible gait. First of all we require the gait to be periodic, so 
\begin{equation}
	\vr{x}_0 = \vr{f}_H(\vr{x}_f),
	\label{eq:heelstrike}
\end{equation}
where $\vr{x}_0$ and $\vr{x}_f$ are the initial and final states of the robot, and $\vr{f}_H$ is the heel-strike map. The states $\vr{x}_0$ and $\vr{x}_f$ are unknown a priori, but constrained by \eqref{eq:heelstrike}, which is the particular form of the boundary constraint \eqref{eq:OCP_boundary} in this case. To construct $\vr{f}_H$ it is assumed that, at heel strike, an impulsive collision occurs that changes the joint velocities but not their angles, and that, as soon as the leading foot impacts the ground, the trailing foot loses contact with it. The collision conserves angular momentum but introduces an instantaneous drop of kinetic energy in the system \citep{kelly2017introduction}. Next, we require the robot to march at a certain speed, which is achieved by setting the final time of the period to $t_f = 0.7$s, and the length $D$ in Fig.~\ref{fig:testcases} to $0.5$m. We also constrain the vertical velocity component of the trailing foot to be positive at $t=0$, and negative when it touches the ground for $t=t_f$. Finally, we require the swing foot to be above the ground at all times.  An animation of the solution we obtain can be seen in
\url{https://youtu.be/dtS-WbESiW0}.

Figure \ref{fig:5L_dyn_err} shows the second order errors $\varepsilon^{[2]}$ for the different collocation methods. As before, the number of intervals used in the trapezoidal cases is twice those used in Hermite-Simpson ones so as to have identical number of collocation points and have balanced comparisons. The results are qualitatively similar to those of the cart-pole, though here the error diminution obtained by the new methods is even more accentuated. As we can see in Table \ref{table:5Lerrors}, the integral second order error $E^{[2]}$ of \HS{2} improves in more than one order of magnitude that of \HS{1} even using a slightly lower computation time. In the case of \TZ{2}, its improvement over \TZ{1} is still higher, reaching a reduction factor near $66$, and using a computation time only slightly longer.

\subsection{A ball throwing problem}
\label{subsec:panda}

\begin{figure*}[t!]
	\begin{center}
		\includegraphics[width=0.49\linewidth]{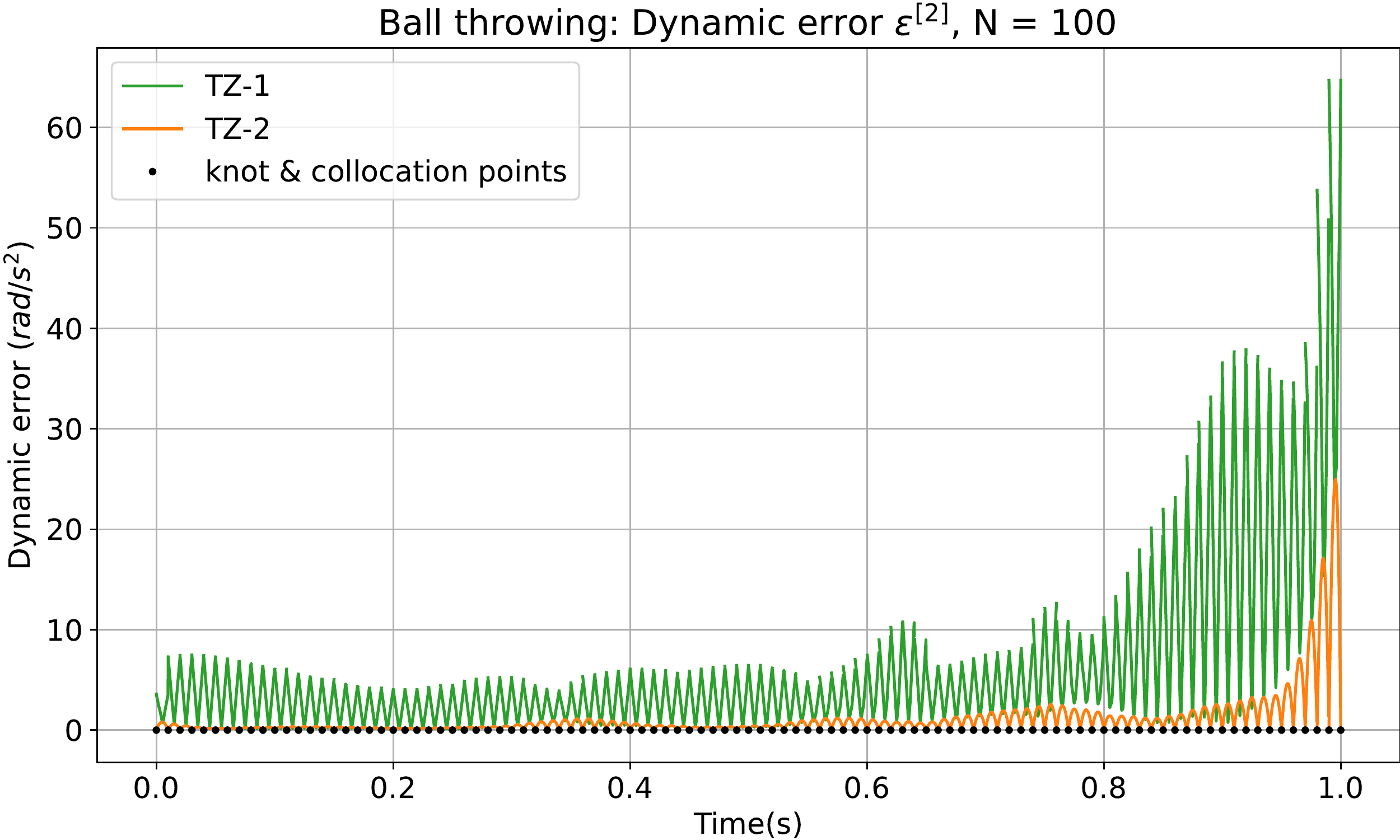}
		\includegraphics[width=0.49\linewidth]{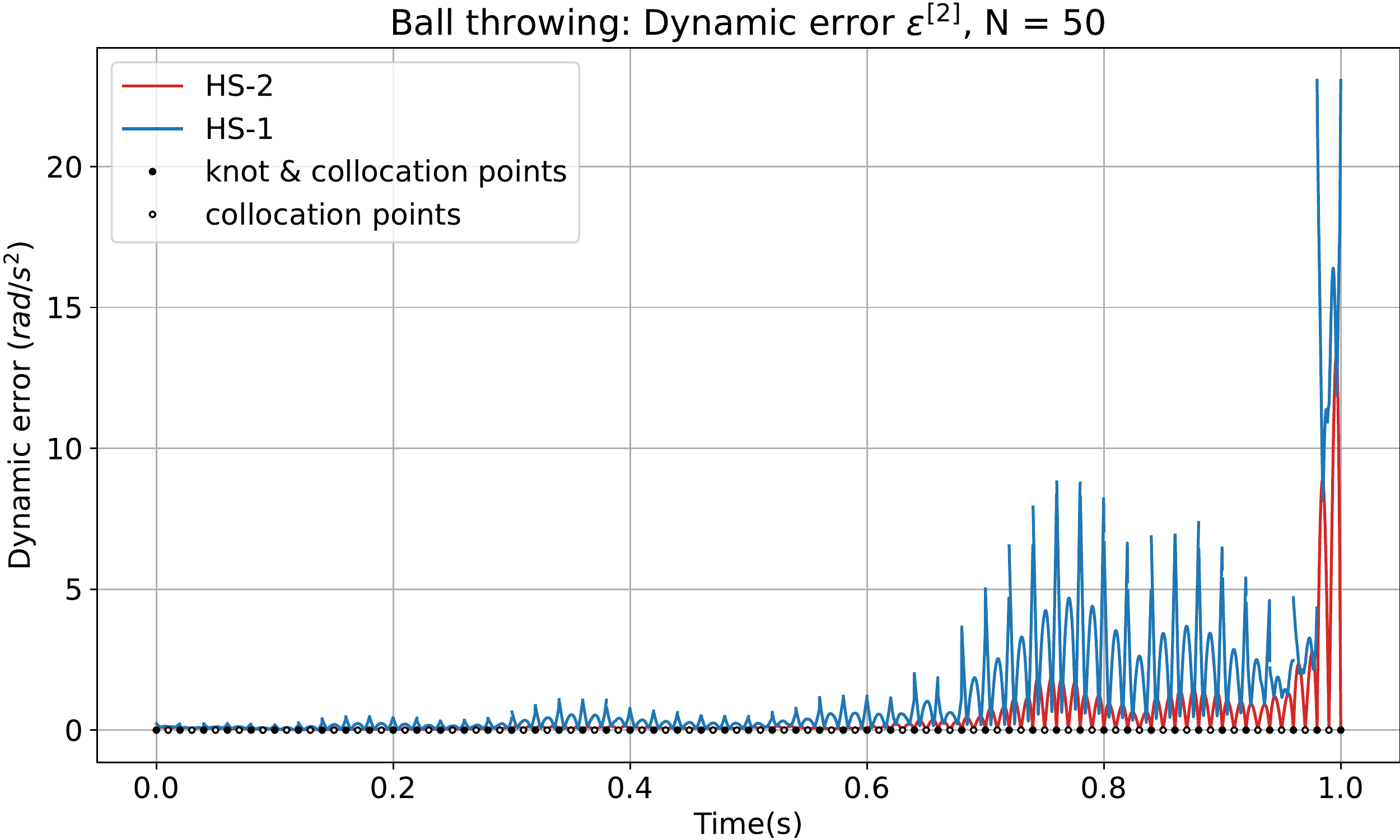}
		\vspace{2mm}
	\end{center}
	\caption{Second order dynamic errors for the ball throwing problem.}
	\label{fig:panda_dyn_err}
\end{figure*}

As a final example, we apply the methods to compute an object-throwing trajectory for a 7R Panda manipulator. The robot is initially at rest, grasping a ball with its gripper, and its task is to throw the ball from the same configuration after $1$ second, with an horizontal velocity of $10$m/s. Since the dynamic model is complex in this case, we rely on the advanced dynamics engine Pinocchio \citep{carpentier2019pinocchio} to compute the $f_k$ and $g_k$ values in the collocation formulas. This engine implements the forward dynamics algorithms by \citet{featherstone2008rigid} in C++, which speeds up the computations considerably. As for the cost function, we use
\begin{equation}  \label{Jpanda}
	\int_{0}^{t_f} 
	\left[ \, \vr{u}(t)\trans\vr{u}(t) 
	+ K_a
	\ddq(t)\trans \ddq(t) \, \right]  dt,
\end{equation}
where $K_a$ is a small value that we fixed to $0.1$ in our runs. While the first term in the integrand penalizes large control torques, the second helps to achieve smoother trajectories for the state.


To compare the methods on an equal footing, in all runs we feed the NLP solver with an initial guess that allows the convergence to a similar solution. This guess is obtained using the \TZ{1} method with $N=25$, initialized with $u_k=0$ for all $k$, and using $x_k$ values that interpolate the initial state, a guessed state for $t=0.5s$, and the final state. The trajectory obtained via $\TZ{1}$ is then used to warm start all methods in the comparisons. As a reference, Fig.~\ref{fig:panda_sequence} shows the trajectory obtained using \HS{2} and $N=100$. Note how the robot performs a circular motion, exploiting gravity to gain momentum, so as to get back to the launch point with the required speed.

Fig.~\ref{fig:panda_dyn_err} compares the dynamic error $\varepsilon^{[2]}$ for the trapezoidal and Hermite-Simpson methods (left and right plots, respectively). As in the previous problems, the new methods notably outperform the conventional ones in terms of this error. The integral errors $E^{[2]}$ corresponding to these figures can be seen in Table~\ref{table:pandaerrors}, together with those of $E^{[1]}$ and $t_c$, confirming similar trends as in the earlier problems.

\subsection{Performance scaling with $N$}
\label{subsec:scaling}

\begin{figure*}[t!]
	\begin{center}
 
		\includegraphics[width=.49\linewidth]{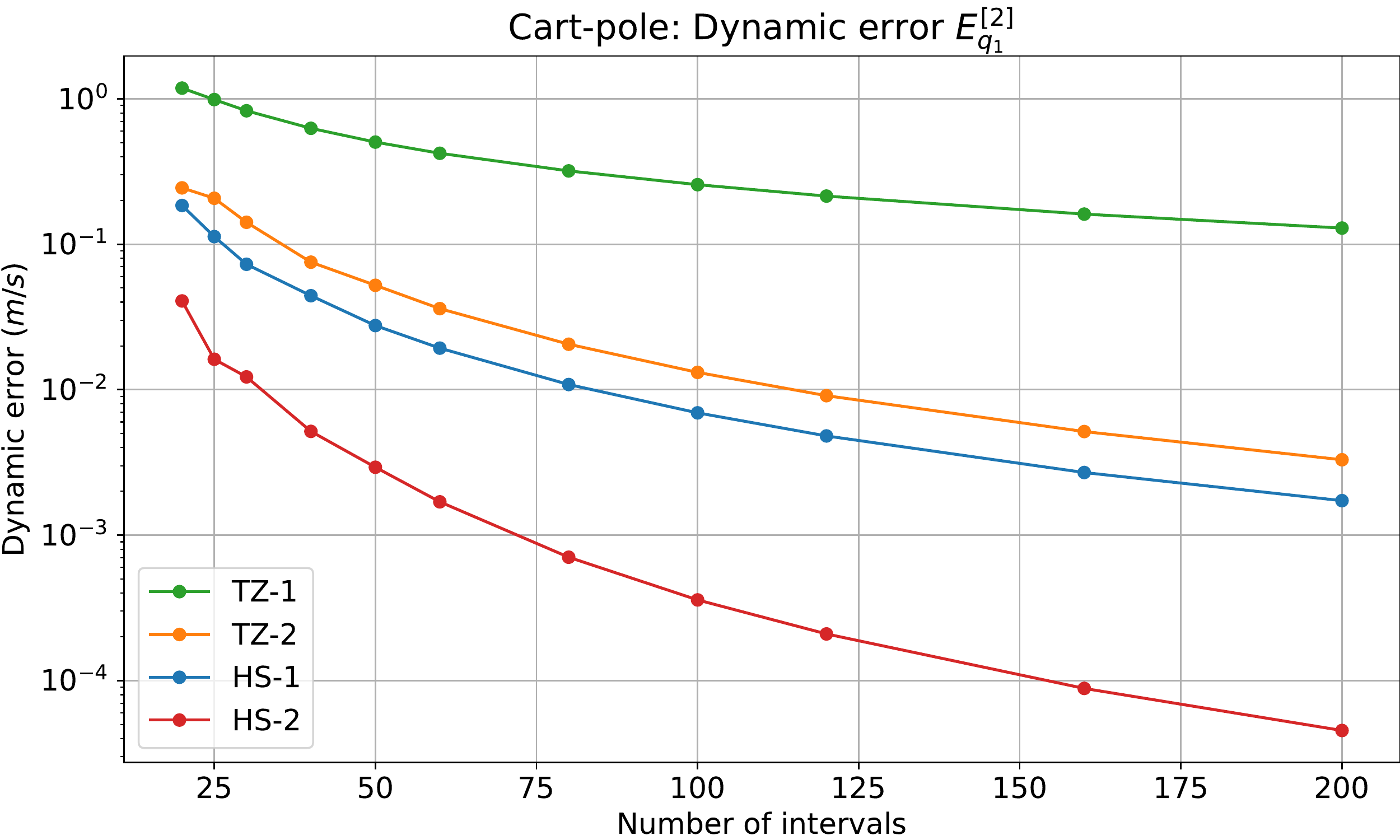}
		\includegraphics[width=.49\linewidth]{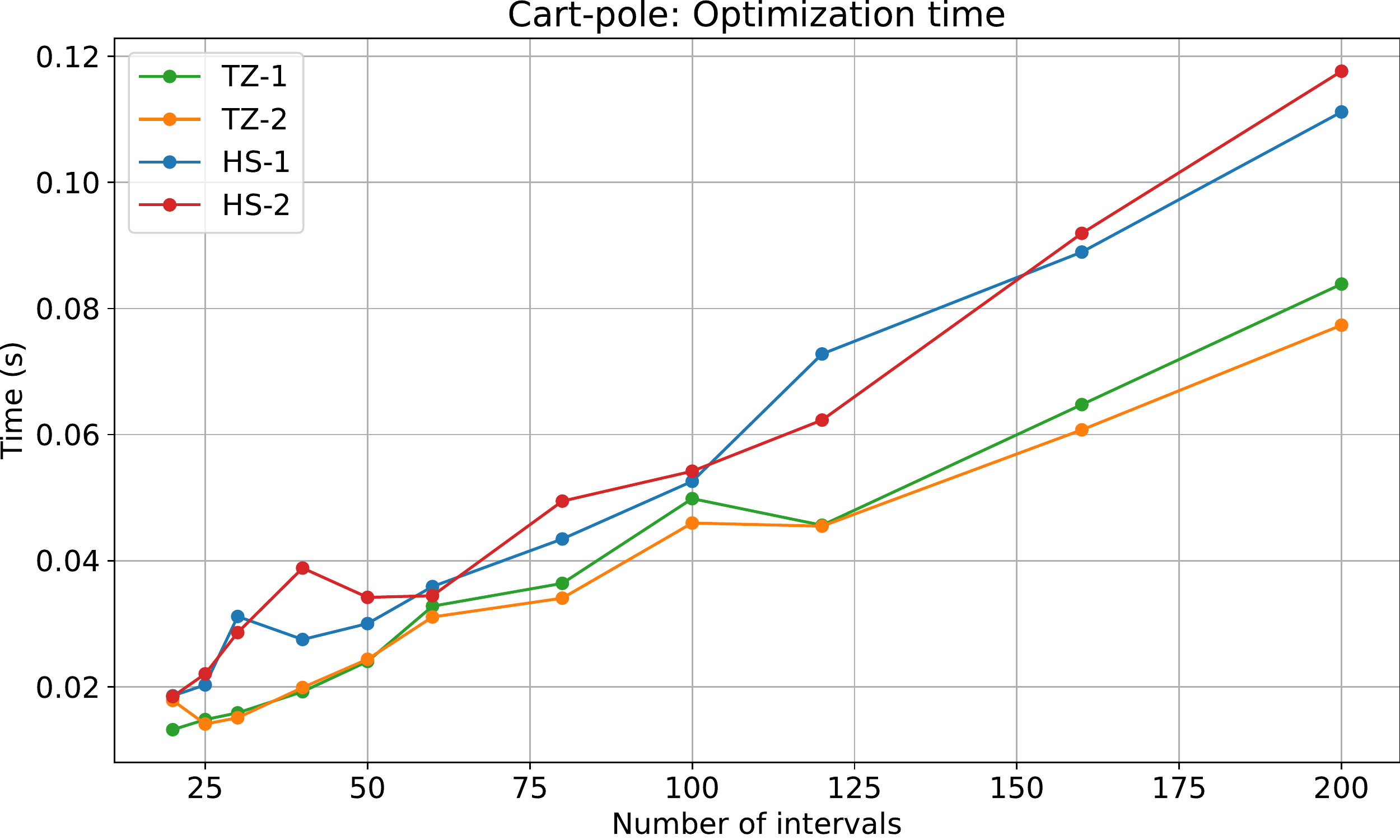}
		
		\vspace{1mm}
		
		\includegraphics[width=0.49\linewidth]{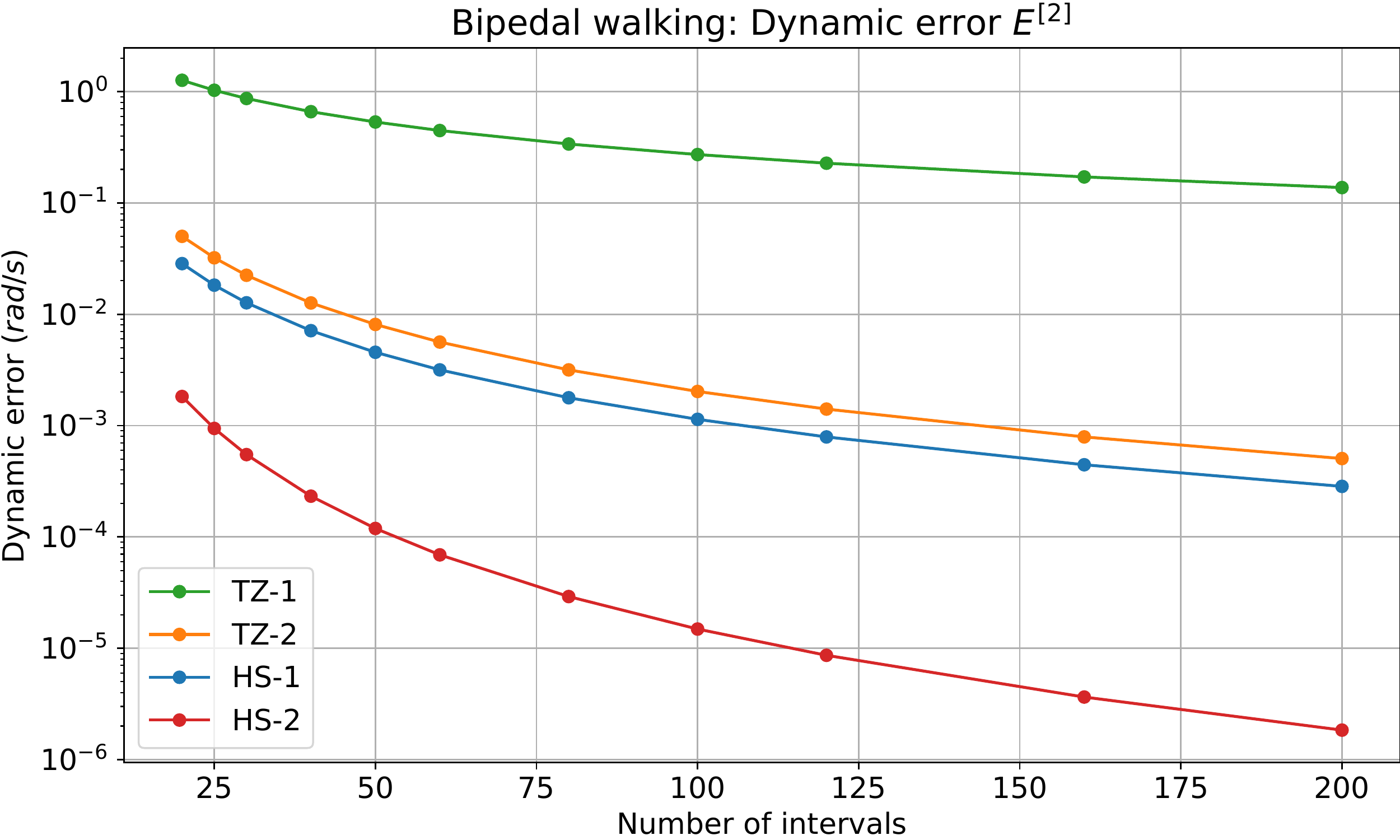}
		\includegraphics[width=0.49\linewidth]{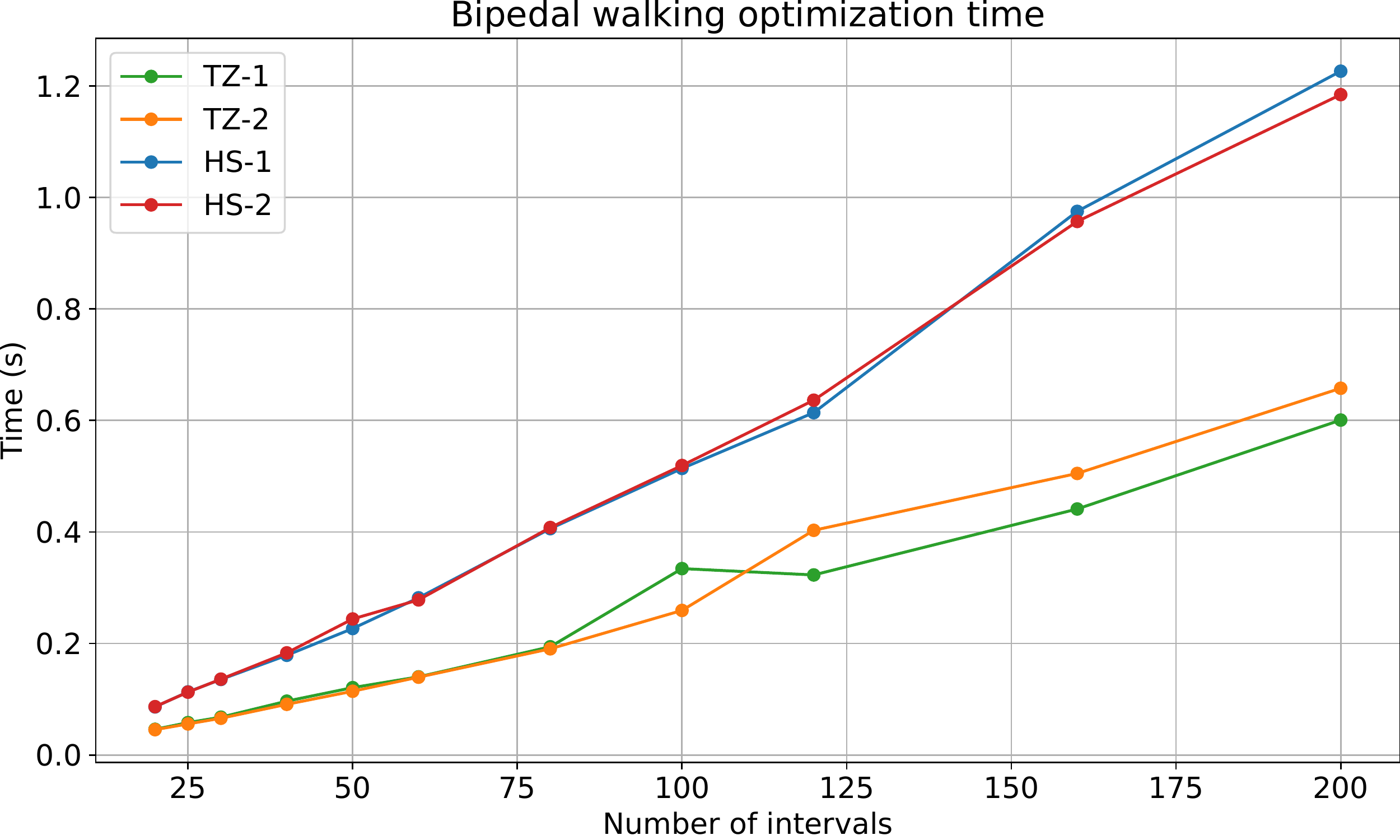}
		
		\vspace{1mm}
		
		\includegraphics[width=0.49\linewidth]{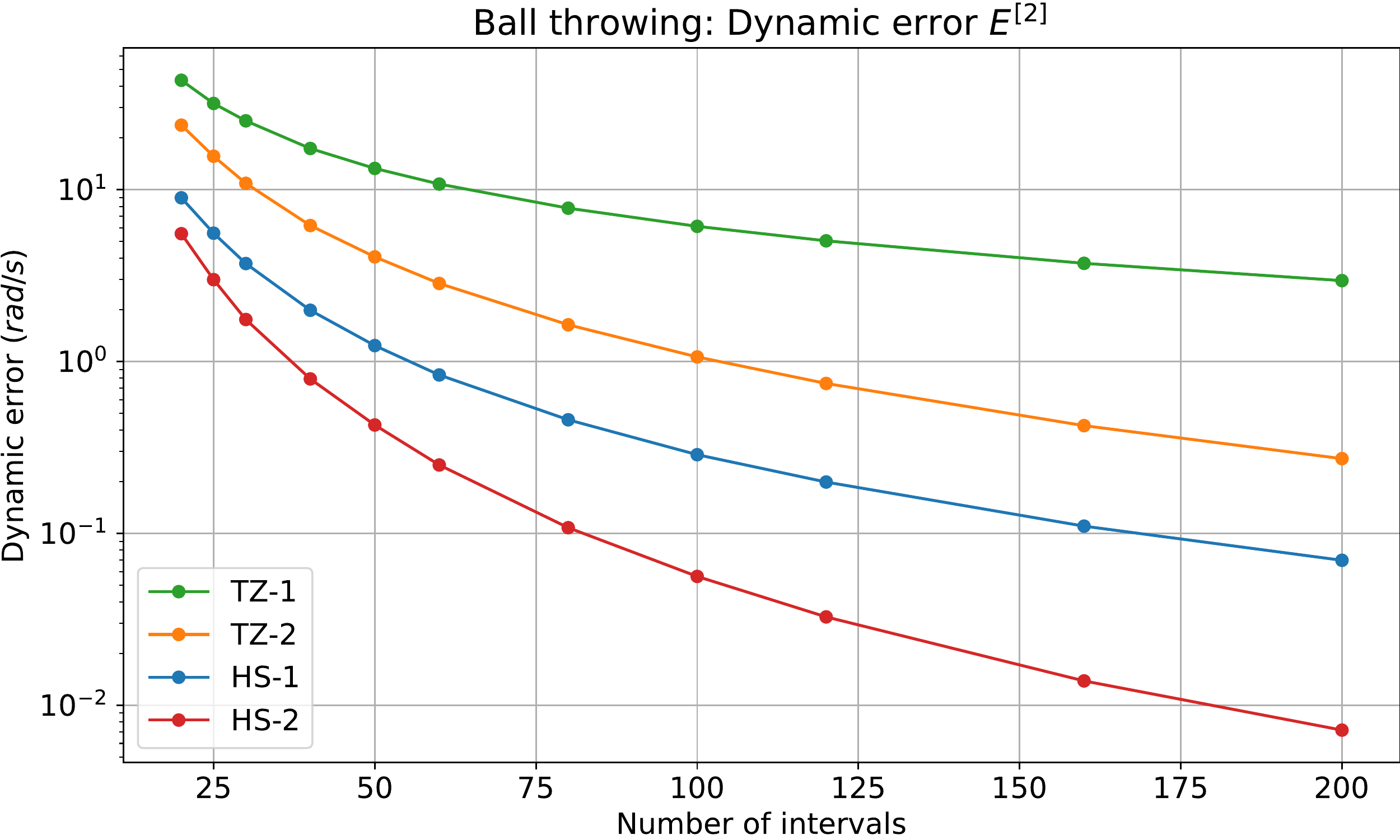}
		\includegraphics[width=0.49\linewidth]{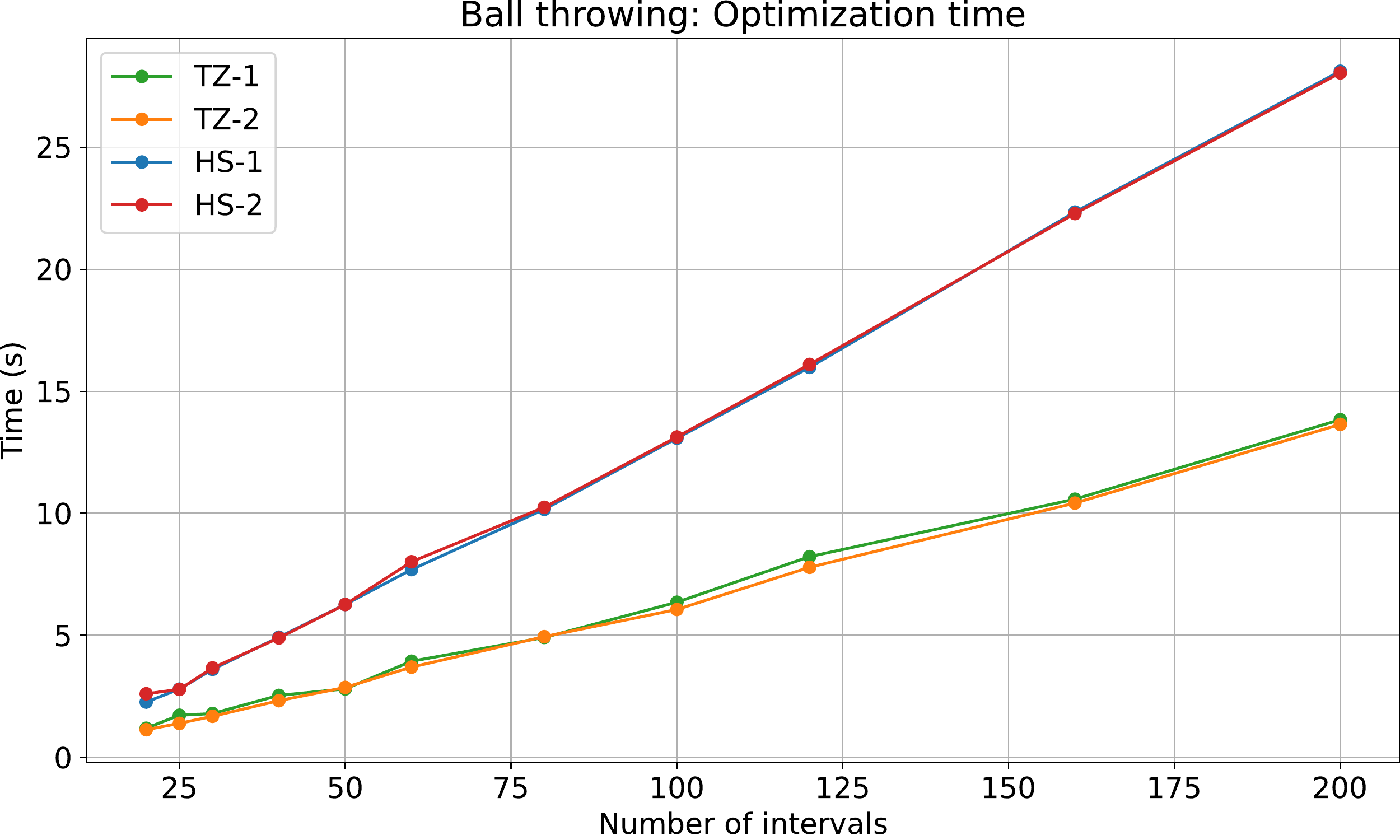}
		
	\end{center}
	\caption{Dynamic error (left) and optimization time (right) for the three test problems, as $N$ is increased.}
	\label{fig:5L_review_graphs}
 \end{figure*}

To evaluate the performance of the methods when the number $N$ of intervals increases, a series of experiments have been conducted by progressively rising $N$ from $20$ to $200$. 
Each experiment has been launched several times and the average of the integral second order errors and computation times are represented in Fig.~\ref{fig:5L_review_graphs}. For the bipedal walking and ball throwing problems we provide $E^{[2]}$. For the cart-pole problem we cannot use $E^{[2]}$ as its coordinates $q_1$ and $q_2$ have different units. Instead we provide only the plot of $E^{[2]}_{q_1}$, since the one of $E^{[2]}_{q_2}$ is very similar.

\begin{table}[t!]
	\begingroup
	\begin{center}
		\setlength{\tabcolsep}{13pt}
		\begin{tabular}{@{}ccccc@{}}
			\toprule
			& $N$& $t_c$ & $E^{[1]}$ & $E^{[2]}$  \\[.3em] 
			&    & $(s)$ & $(rad)$ & $(rad/s)$\\
			\midrule
            \TZ{1} & 100 & 2.487 & 0.0189  & 6.1183 \\
			\TZ{2} & 100 & 2.325 & 0       & 1.0635 \\
		  \HS{1} & 50 & 2.715 & 0.0072  & 2.4398  \\
		  \HS{2} & 50 & 2.451 & 0       & 0.7146  \\ \bottomrule 
        \end{tabular}
	\end{center}
    \vspace{1mm}
	\caption{Performance data for ball throwing problem ($E^{[2]}$ is the integral of $\varepsilon^{[2]}$ in Fig. \ref{fig:panda_dyn_err}).
	\label{table:pandaerrors}}
	\endgroup
\end{table}

In the three test problems, the best results for the second order error (shown on the left hand side of Fig.~\ref{fig:5L_review_graphs} in logarithmic scale) are those of \HS{2}, which, in many cases, improve the results of \HS{1} in about one order of magnitude, or even more, and the improvement rate tends to increase with the number $N$ of intervals. The same behavior is observed for \TZ{2} with respect to \TZ{1}. Interestingly, in all cases the performance of \TZ{2} produces, for the same number of intervals $N$, only about twice the error of \HS{1}, and this rate is kept rather constant with $N$. However, a more balanced comparison would be to look at experiments with equal number of collocation points, what means to compare each $N$ value of \HS{1} with the $2N$ value of \TZ{2}. A close look at the plots will convince the reader that this comparison gives equal or better results for \TZ{2} in all cases. 

The right hand side plots in Fig.~\ref{fig:5L_review_graphs} show the growth of the computation times with the number $N$ of intervals. The plots consistently show that the difference in computation time between a method for first order systems and the corresponding method for second order systems is not relevant. In all cases, the growth is nearly linear in $N$, but the increase rate is higher for the \HS{} methods than for the \TZ{} ones. Despite the different complexity of the three problems analyzed, reflected in the different time scales involved, in all cases, the increase rate of the \HS{} methods is nearly twice that of the \TZ{} methods. In other words, the increasing rate is very similar for all methods when comparing the computation times for the same number of collocation points.

\section{Conclusions}
\label{sec:conclusions}

Trapezoidal and Hermite-Simpson collocation methods are very popular in the robotics community. However, they are conceived for dynamical systems of first order, while the dynamics of the systems found in robotics are often $M$th order, with $M > 1$. The transcription of an $M$th order ODE as a first order one has the unexpected effect that the dynamic equations are not actually imposed at the collocation points. Properly imposing the $M$th order constraints at the same such points as in the original algorithms requires increasing the degree of the polynomials approximating the configuration trajectory, while keeping the implied degrees for its time derivatives. This is achieved with the methods we propose, which grant the functional consistency between the trajectories of all the state coordinates, not only at the collocation points, but also along the computed trajectory. Using benchmark problems of increasing complexity, we have also shown that the new methods provide trajectories with a much smaller dynamic error than those of conventional methods, despite they require a comparable amount of computation time. This implies that the obtained trajectories will be more compliant with the system dynamics, so they should be easier to track with a feedback controller. Moreover, the  trajectories of the new methods are $M$ times differentiable, so in addition to enjoying smooth velocities, their accelerations will be continuous, or even the jerk if $M\geq3$, which are very desirable properties from a control perspective.

Points that deserve further attention are the extension of these ideas to pseudospectral collocation methods, which we initially explored for \mbox{$M=2$} in \cite{moreno2022legendre}, or generalizations to deal with constrained multibody systems \citep{posa2016optimization,bordalba2023direct}, floating-base systems \citep[Chapter 17]{tedrake2023underactuated}, or limits on jerk or higher-order derivatives of the trajectories.

\backmatter

%

\bmhead{Acknowledgments}

This work has been partially funded by Agencia Estatal de 
Investigaci\'on under project {\em Kinodyn+}, with reference 
PID2020-117509GB-I00 / AEI / 10.13039/50110001103. 


{
	\small
    \balance
	\bibliographystyle{apalike}
	\bibliography{references}
}

\end{document}